\documentclass[10pt,twocolumn,letterpaper]{article}

\usepackage{cvpr}              

\usepackage{multirow}

\usepackage{algorithm}
\usepackage{soul}
\usepackage{xcolor}
\usepackage{multirow}
\usepackage{bbding}
\usepackage{float} 
\usepackage{subcaption}
\usepackage{amssymb}
\usepackage{amsmath}

\usepackage{times}
\usepackage{helvet}
\usepackage{courier}
\usepackage{xcolor}

\usepackage{amsmath}
\usepackage{algorithm}
\usepackage{algpseudocode}
\usepackage{placeins}
\usepackage[accsupp]{axessibility}

\algnewcommand\Input{\item[\textbf{Input:}]}%
\algnewcommand\Output{\item[\textbf{Output:}]}%
\definecolor{cvprblue}{rgb}{0.21,0.49,0.74}
\usepackage[pagebackref,breaklinks,colorlinks,allcolors=cvprblue]{hyperref}
\newcommand{\model}{CogDriver} 

\def\paperID{*} 
\def\confName{CVPR}
\def\confYear{2026}

\title{CogDriver: Integrating Cognitive Inertia for Temporally Coherent Planning in Autonomous Driving}

\author{Pei Liu$^{1,2*}$\quad  Qingtian Ning$^{3*}$\quad Xinyan Lu$^{3}$\quad Haipeng Liu$^{3}$\quad Weiliang Ma$^{3\dagger}$\quad Dangen She$^{3}$\quad \\Xianpeng Lang$^{3}$\quad Jun Ma$^{1,2\ddagger}$\\
$^{1}$The Hong Kong University of Science and Technology (Guangzhou) \\ $^{2}$The Hong Kong University of Science and Technology \quad $^{3}$Li Auto Inc.\\
{\tt\small pliu061@connect.hkust-gz.edu.cn, jun.ma@ust.hk}
}

\begin{document}
\maketitle
\let\thefootnote\relax\footnotetext{$^*$Equal contribution. $^\dagger$Project leader. $^\ddagger$Corresponding author.}
\begin{abstract}

The pursuit of autonomous agents capable of temporally coherent planning is hindered by a fundamental flaw in current vision-language models (VLMs): they lack cognitive inertia. Operating on isolated snapshots, these models cannot form a continuous understanding of the environment, leading to erratic decision jitter and a failure to execute complex, multi-step maneuvers. To remedy this, we introduce CogDriver, a framework designed to build a stable internal representation by instilling this crucial cognitive property. Our work makes two key contributions: (1) We present CogDriver-Data, a large-scale vision-language-action dataset whose narrative annotations provide the supervisory signal for learning temporal dynamics and persistent intent. (2) We develop the CogDriver-Agent, an architecture featuring a sparse temporal memory to maintain a stable internal state. This is enabled by a spatiotemporal knowledge distillation approach that explicitly teaches decision coherence. Comprehensive experiments validate our paradigm: CogDriver-Agent achieves a 22\% increase in the closed-loop Driving Score on Bench2Drive and a 21\% reduction in mean L2 error on nuScenes, establishing a new state-of-the-art. These significant gains in both long-term decision-making and imitation accuracy provide strong evidence that our agent successfully maintains a temporally coherent internal state, bridging the gap toward more reliable autonomous driving. Project link: \href{https://ocean-luna.github.io/CogDriver.github.io/}{CogDriver}.

\end{abstract}
\begin{figure*}

    \centering
    \includegraphics[width=0.8\linewidth]{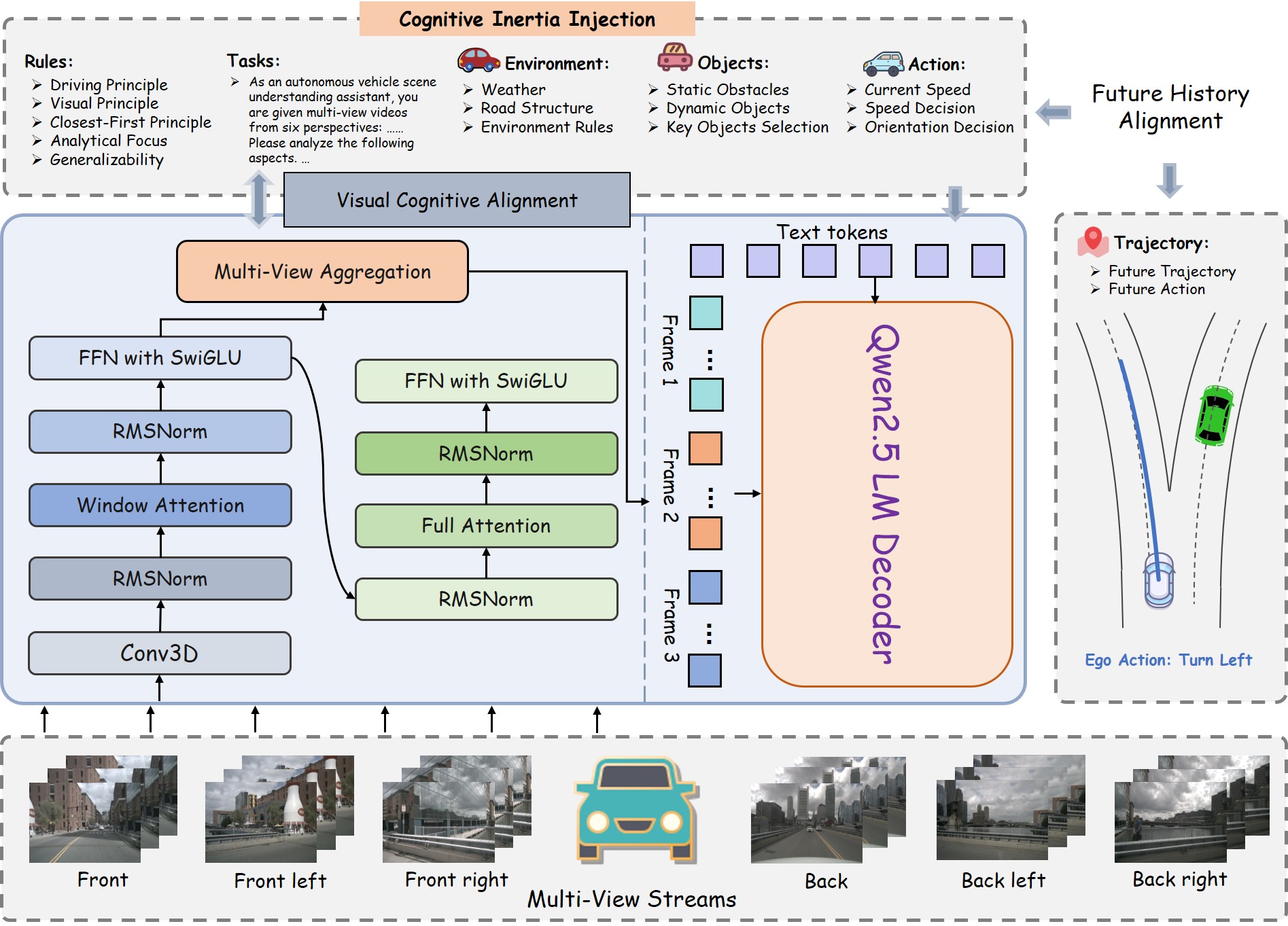}
    \caption{Annotation pipeline. To generate temporally coherent data, we propose a novel Multi-View Spatiotemporal MLLM capable of processing concurrent video streams. Its reasoning is guided by our cognitive inertia Injection framework, which provides structured rules and tasks. The generated narrative is then rigorously verified against the ground-truth vehicle trajectory via Future History Alignment, ensuring the final annotations are both causally sound and coherent.}
    \label{fig:annotation}
\end{figure*}

\section{Introduction}
The grand challenge of autonomous driving is not merely to create a system that perceives, but one that understands and anticipates the world through stable, temporally coherent reasoning \cite{ge2024worldgpt, yildirim2024task, zhao2025world}. While much of the autonomous driving literature focuses on "world models" that explicitly predict future states in pixel or latent space, a foundational prerequisite for such anticipation is the ability to maintain a consistent internal representation over time. However, a fundamental and often overlooked flaw plagues current end-to-end systems: they operate as cognitively stateless agents, trapped in a perpetual present \cite{zhou2024vision, chen2024vadv2}. Lacking a continuous thread of thought, they are incapable of building a stable internal state; instead, they process each moment as an isolated problem, akin to an amnesiac driver re-evaluating the world from scratch every fraction of a second.

The advent of large vision-language models (VLMs) presents a tantalizing opportunity to imbue driving systems with advanced reasoning capabilities \cite{liu2024survey, zhou2024gpt}. Yet, naively applying them often exacerbates the problem. Current VLMs suffer from a lack of cognitive inertia, the natural persistence of intentions that underpins stable human cognition \cite{radford2021learning, zhai2023sigmoid}. This deficit stems from a more fundamental failure: the inability to maintain temporal coherence. Without this cognitive anchor, an agent's internal representation is fragmented and ephemeral. This manifests externally as debilitating decision jitter and a failure to execute complex, multi-step maneuvers. Consider, for instance, the complex decision of overtaking a slow-moving truck on a multi-lane highway. An agent without cognitive inertia might initially decide to overtake from the left. However, upon detecting a fast-approaching vehicle in the left lane for a brief moment, it might abruptly cancel the maneuver and swerve back, only to then reconsider overtaking from the right as that lane momentarily clears. This erratic oscillation between competing strategies, "overtake left," "abort," "overtake right", makes the vehicle dangerously unpredictable. In contrast, a human driver forms a stable, multi-step plan, such as: "The left lane is too risky; I will wait for the car on the right to pass, then execute a clean overtake from the right." This commitment to a chosen strategy, guided by a coherent understanding of the environment, is not a luxury; it is a fundamental prerequisite for safe and trustworthy interaction in a dynamic world.

This leads to a more fundamental research question: \textbf{How can we build VLA agents that develop a coherent internal representation, enabling them to act with the stability and foresight of a human?}

To answer this, we introduce {\model}, a framework designed not merely to process temporal data, but to lay the groundwork for temporally coherent planning by instilling cognitive inertia. Our work pioneers this direction through two key contributions. First, we present {\model}-Data, two large-scale vision-language-action (VLA) datasets. Unlike existing datasets with disconnected rationales, {\model}-Data's narrative annotations capture the story of persistent driving intent. This narrative structure provides the crucial supervisory signal for learning the temporal dynamics essential for stable planning.

Second, building on this foundation, we develop the {\model}-Agent. Its architecture is designed to maintain a stable internal state, a prerequisite for consistent decision-making. It features a sparse temporal memory module that acts as a mechanism for intentional persistence, effectively forming a coherent temporal context over time. This is enabled by our spatiotemporal knowledge distillation approach, which explicitly teaches the model to maintain decision coherence by learning from the narrative structure of {\model}-Data.

Comprehensive experiments show that {\model}-Agent not only achieves state-of-the-art performance but, more critically, exhibits significantly reduced decision jitter and successfully executes long-term plans. These emergent capabilities serve as strong evidence that our agent is developing a more stable internal representation. This work represents a paradigm shift from building reactive predictors to engineering cognitively coherent agents, marking a crucial step towards trustworthy autonomous systems powered by robust temporal reasoning.

Our contributions are summarized as follows:

\begin{itemize}
\item We present {\model}-Data, two large-scale VLA datasets with novel narrative annotations designed to capture persistent intent and temporal coherence, providing the first benchmark for training and evaluating cognitively coherent agents.

\item We propose {\model}-Agent, an architecture designed to build a coherent internal representation by instilling cognitive inertia. It features a sparse temporal memory to maintain a stable internal state, trained via a knowledge distillation method that promotes decision coherence.

\item We demonstrate through extensive experiments that our approach not only sets a new state-of-the-art in standard benchmarks but also quantifiably reduces decision jitter and enables the execution of complex, long-term plans, validating the effectiveness of our paradigm.
\end{itemize}

\begin{table*}[]
    \centering
    \renewcommand\arraystretch{1.3}
     \caption{Comparison of E2E autonomous driving datasets with language and action data. M.V.: Multi-View images; Auto.: Auto-labeling; Trfc.: Traffic; Sur.: Surrounding; Obj.: Object; Pos.: Position; Traj.: Trajectory.}
    \scalebox{0.75}{
    \begin{tabular}{c|ccccc| ccc ccccccc}
    \hline
    \hline
    \multirow{2}{*}{\textbf{Dataset}} & \multicolumn{2}{c}{\textbf{Vision Data}} & \multicolumn{2}{c}{\textbf{Language Data}} & \multicolumn{1}{c|}{\textbf{Action Data}} &  Reason & Cogni & Weat & Road & Trfc. & Trfc. & Sur. & Obj.\\
    ~ & M.V. & Temporal & VQA & Auto. &  Type  & -ing & -tive & -her & Type  & Light & Sign & Obj. & Pos.\\
    \hline
    
         \multicolumn{1}{l|}{Talk2Car \cite{deruyttere2019talk2car}} & \XSolidBrush & \XSolidBrush &\XSolidBrush &\XSolidBrush  &Traj. & \XSolidBrush  & \XSolidBrush & \XSolidBrush & \XSolidBrush & \XSolidBrush & \XSolidBrush & \Checkmark & \XSolidBrush\\
         
         \multicolumn{1}{l|}{T2C-Traj \cite{deruyttere2022talk2car}} &\XSolidBrush  & \XSolidBrush  & \XSolidBrush &\XSolidBrush &Traj., Command&  \XSolidBrush & \XSolidBrush & \XSolidBrush & \XSolidBrush & \XSolidBrush & \XSolidBrush & \Checkmark & \XSolidBrush\\
         
         \multicolumn{1}{l|}{DriveLM-nuScenes \cite{sima2024drivelm}} & \XSolidBrush & \XSolidBrush& \Checkmark &\XSolidBrush  &Traj. & \Checkmark & \XSolidBrush & \XSolidBrush & \XSolidBrush & \XSolidBrush & \XSolidBrush & \Checkmark & \XSolidBrush\\
         \multicolumn{1}{l|}{DriveLM-CARLA \cite{sima2024drivelm}} & \Checkmark & \XSolidBrush& \Checkmark& \Checkmark  &Traj. & \Checkmark & \XSolidBrush & \XSolidBrush & \XSolidBrush & \XSolidBrush & \XSolidBrush & \Checkmark & \XSolidBrush\\
         
         \multicolumn{1}{l|}{DRAMA \cite{malla2023drama}} &  \XSolidBrush & \Checkmark &  \Checkmark & \Checkmark &  Command & \Checkmark & \XSolidBrush & \XSolidBrush & \Checkmark & \Checkmark & \Checkmark & \Checkmark &\XSolidBrush\\
         \multicolumn{1}{l|}{Rank2Tell \cite{sachdeva2024rank2tell}}& \XSolidBrush & \Checkmark & \XSolidBrush & \Checkmark & Command & \Checkmark & \XSolidBrush & \XSolidBrush & \XSolidBrush & \Checkmark & \XSolidBrush & \Checkmark & \Checkmark\\
         
         \multicolumn{1}{l|}{Reason2Drive \cite{nie2024reason2drive}}& \XSolidBrush & \XSolidBrush & \Checkmark & \Checkmark  & Command & \Checkmark & \XSolidBrush & \XSolidBrush & \XSolidBrush & \XSolidBrush & \XSolidBrush & \Checkmark & \XSolidBrush\\

         \multicolumn{1}{l|}{CoVLA \cite{arai2025covla}} & \XSolidBrush & \XSolidBrush&\XSolidBrush& \Checkmark& Traj.& \XSolidBrush & \XSolidBrush & \XSolidBrush & \Checkmark & \Checkmark & \XSolidBrush & \Checkmark & \XSolidBrush\\
         
         \multicolumn{1}{l|}{Omnidrive \cite{wang2025omnidrive}} & \Checkmark & \XSolidBrush  & \Checkmark & \Checkmark  & Traj. &\Checkmark & \Checkmark & \Checkmark & \XSolidBrush & \Checkmark & \Checkmark & \Checkmark & \Checkmark\\
         
        \hline
         
         \multicolumn{1}{l|}{{\model}-nuScenes (Ours)} & \Checkmark & \Checkmark & \Checkmark & \Checkmark & Traj., Command & \Checkmark &\Checkmark &\Checkmark & \Checkmark & \Checkmark &\Checkmark & \Checkmark & \Checkmark\\
         \multicolumn{1}{l|}{{\model}-Bench2Drive (Ours)} & \Checkmark & \Checkmark & \Checkmark & \Checkmark & Traj., Command & \Checkmark &\Checkmark &\Checkmark & \Checkmark & \Checkmark &\Checkmark & \Checkmark & \Checkmark\\
         \hline
         \hline
    \end{tabular}}
    \label{tab:data}
\end{table*}

\section{Related Work}
\subsection{Language-enhanced Dataset for Autonomous Driving}
The evolution of driving datasets has recently shifted towards incorporating language to explain actions \cite{li2023open, liu2024survey}. However, existing works fundamentally fail to capture cognitive inertia, treating driving as a series of disconnected moments. Pioneering datasets like BDD-X \cite{kim2018textual} and DRAMA \cite{malla2023drama} offer only atomic, snapshot-based rationales. While subsequent works like DriveLM \cite{sima2024drivelm} and CoVLA \cite{arai2025covla} introduced continuous trajectories, they still lack the corresponding "continuous why", the evolving causal narrative that connects decisions over time. Thus, no existing dataset provides the essential ingredients to learn cognitive inertia: the joint modeling of continuous action, holistic multi-view perception, and a persistent, evolving thought process. A thorough comparison between existing prompt-based driving datasets and ours is summarized in Table \ref{tab:data}.



\subsection{MLLMs Grounded Autonomous Driving}

Recent multimodal large language models (MLLMs) for driving, despite diverse architectures like prompt-based planners \cite{mao2023gpt, chen2024driving}, end-to-end frameworks \cite{xu2024drivegpt4}, and hybrid designs \cite{shao2024lmdrive, wang2023drivemlm, renz2024carllava}, largely operate as stimulus-response mechanisms. They fundamentally fail to model the causal relationships within dynamic scenes, a critical capability for explainable decision-making. Our work addresses this gap by proposing a framework that explicitly reasons over spatiotemporal causal dependencies. This approach unlocks zero-shot E2E planning, advancing beyond the conventional stimulus-response mappings of prior studies \cite{han2025dme, mao2023language, marcu2312lingoqa, pan2024vlp, seff2023motionlm, yuan2024rag} by grounding decisions in causal-temporal principles.

\begin{figure*}
    \centering
    \includegraphics[width=0.9\linewidth]{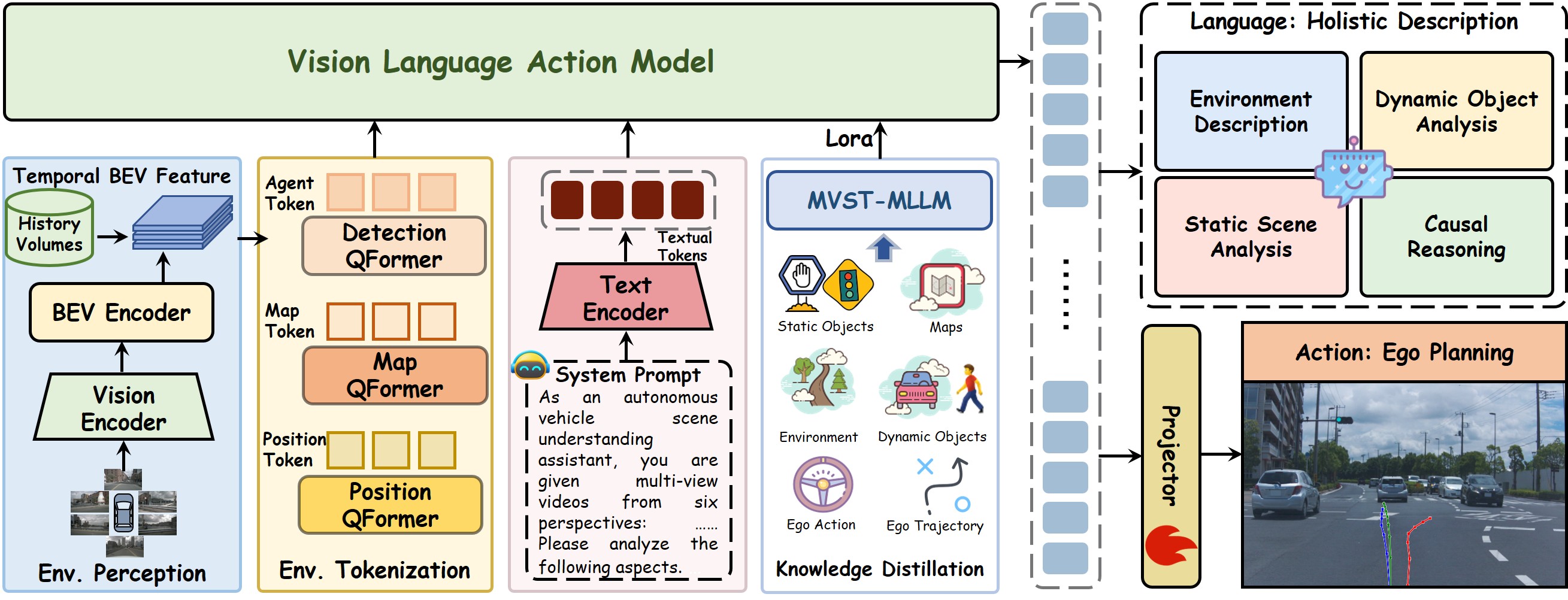}
    \caption{An overview of our {\model}-Agent. It moves beyond reactive decision-making by harnessing a pre-trained language model to maintain cognitive inertia. It achieves this by building a stable internal world model that continuously integrates 3D perception, ego states, and language commands. This allows the model to generate not just context-aware, but temporally coherent plans. The model's effectiveness is demonstrated by its state-of-the-art performance across both open-loop trajectory planning and complex closed-loop driving tasks. Its success highlights a unique capability: bridging perception with action through a persistent, evolving strategy, rather than disconnected, stimulus-response mappings.}
    \label{fig:model}
\end{figure*}

\section{{\model}-Data}

We introduce {\model}-Data, a new suite of large-scale VLA datasets, {\model}-nuScenes, and {\model}-Bench2Drive, designed to address a critical gap in existing autonomous driving data: the lack of narrative coherence. While prior datasets provide per-frame rationales, {\model}-Data features annotations that tell a continuous story of driving decisions, capturing persistent intent, causal reasoning, and the resulting actions. As illustrated in Fig. \ref{fig:annotation}, these narrative-rich annotations are generated by a novel pipeline designed to instill spatiotemporal and logical consistency, providing a robust foundation for training cognitively coherent agents.

\subsection{Data Annotation Pipeline}

Our annotation pipeline is built upon two core technical innovations designed to generate data with unprecedented spatiotemporal richness and logical consistency.

First, we propose a novel Multi-View Spatiotemporal MLLM (MVST-MLLM) architecture as the annotator. Unlike prior works that process static images or single video streams, our model's vision encoder is the first of its kind to process multi-view streams concurrently. It employs a hierarchy of Conv3D and window attention layers to extract and fuse features across both space (all camera views) and time. This holistic perception is critical, enabling the model to reason about complex dynamic events that are only comprehensible by correlating information from multiple viewpoints simultaneously, e.g., a car merging from the right while a pedestrian appears on the left.

Second, we introduce cognitive Inertia injection, a specific mechanism that enforces temporal coherence. This is not a vague concept but a direct alignment of static rules with dynamic visual evidence. The MLLM is conditioned on a structured prompt containing high-level Rules and Tasks. The core innovation lies in training the model to use these static principles to generate a single, continuous narrative that explains the entire temporal sequence of visual inputs. This alignment compels the model to produce causally-linked explanations rather than disconnected, per-frame descriptions. This process is further grounded by Future History Alignment, which verifies the generated narrative against the actual vehicle trajectory, ensuring physical plausibility.

The resulting {\model}-Data is therefore a direct product of this architecture's unique ability to see in 360 degrees over time and its training to bind static rules to dynamic events, creating a robust foundation for our agent. Further details are provided in the Appendix.

\section{{\model}-Agent}
We present {\model}-Agent, an end-to-end framework engineered to instill cognitive inertia by seamlessly integrating spatiotemporal perception with language-conditioned planning. As depicted in Fig. \ref{fig:model}, the agent's architecture is not a simple pipeline but a synergistic system. Its perceptual foundation is a hierarchical vision backbone that distills raw multi-view inputs into a compact set of spatiotemporal tokens. Central to this process is our Temporal Coherence Module (TCM), which maintains a dynamic, memory-efficient representation of the world state, enabling robust long-range reasoning. These distilled world tokens are then projected into the embedding space of a frozen VLM. Here, they are fused with historical state information and natural language instructions, allowing a set of lightweight, trainable adapters to steer the powerful, pre-trained VLM core towards generating coherent and actionable trajectories.

\subsection{Temporal Coherence Module}

\begin{algorithm}[t]
\caption{Temporal Coherence Module}
\label{alg:temporal_module_compact}
\begin{algorithmic}[1]
\Require Historical queries $Q_{hist\_c}, Q_{hist\_p}$; Current queries $Q_{init\_c}$; Image features $F_{img\_t}$; Ego-motion $E_{ego}, v, \Delta t$.
\Ensure Refined queries for current frame $Q_{refined\_t}$.

\State \Comment{\textit{1. Motion-Aware Propagation}}
\State $Q_{aligned\_p} \gets E_{ego} \cdot Q_{hist\_p}$ \Comment{Align historical positions}
\State $\alpha, \beta \gets \text{MotionEncoder}(E_{ego}, v, \Delta t)$ \Comment{Compute affine parameters}
\State $Q_{pe} \gets \alpha \cdot \text{LN}(\psi(Q_{aligned\_p})) + \beta$ \Comment{Modulate positional embeddings}
\State $Q_m \gets \alpha \cdot \text{LN}(Q_{hist\_c}) + \beta$ \Comment{Propagate context queries}

\Statex
\State \Comment{\textit{2. Hybrid Attention and Fusion}}
\State $Q_{hybrid} \gets \text{Concat}(Q_m, Q_{init\_c})$
\State $Q'_{hybrid} \gets \text{SelfAttention}(Q_{hybrid})$
\State $Q_{refined\_t} \gets \text{CrossAttention}(Q'_{hybrid}, F_{img\_t} + Q_{pe}, F_{img\_t})$

\State \textbf{return} $Q_{refined\_t}$
\end{algorithmic}
\end{algorithm}

To maintain a coherent world model over time, the agent must solve the critical challenge of tracking object states despite ego-motion and occlusions. Our TCM, detailed in Alg. \ref{alg:temporal_module_compact}, addresses this through an elegant, three-stage process of geometric propagation, motion-conditioned state refinement, and evidence-based fusion.

\textbf{Geometric Propagation}. The process begins by explicitly compensating for the vehicle's own movement. Historical 3D object queries $Q_{hist\_p}$ are geometrically warped into the current frame's coordinate system using the ego-motion transformation $\mathcal{E}_{ego}$. This provides a crucial geometric prior, ensuring that the initial state estimate for each object is grounded in physical reality before any feature-level processing occurs.

\textbf{Motion-Conditioned State Refinement}. Simple geometric alignment is insufficient to capture complex object dynamics or perspective shifts. We therefore introduce a novel motion-aware normalization scheme that dynamically refines the propagated state. Instead of using static normalization parameters, we parameterize conditional affine transformation coefficients, $\alpha$ and $\beta$, as a function of the full motion context 
$({E}_{ego},v,\Delta t)$. These coefficients then perform a motion-conditioned modulation on both the positional embeddings $Q_{pe}$ and the propagated context features $Q_m$. This allows the network to learn a sophisticated, feature-level compensation, for instance, by amplifying features for fast-moving objects or down-weighting features for objects that are likely occluded.

\textbf{State Reconciliation and Fusion}. The refined memory queries, $Q_m$, now representing a strong temporal prior, are concatenated with new perception queries, 
$Q_{init\_c}$, which represents fresh evidence from the current frame. A self-attention mechanism then performs state reconciliation, allowing the model to weigh historical beliefs against new observations, update existing tracks, and suppress redundant or spurious detections. Finally, these reconciled queries are grounded back into the current visual evidence through cross-attention. By injecting the modulated positional embeddings 
$Q_{pe}$ into the keys of the image features, we provide explicit spatial guidance, enabling the model to precisely localize and update the state of each object.

This unified design achieves robust object permanence and temporally consistent perception, forming the bedrock upon which the agent's cognitive inertia is built.

\subsection{Training Objectives}

The proposed model employs a composite loss function to jointly optimize 3D object detection and structured scene understanding. The detection objective combines categorical recognition and spatial localization through two key terms: the classification loss $\mathcal{L}_{cls}$ for object categories is formulated via Focal Loss, while the regression loss $\mathcal{L}_{reg}$ for 3D bounding box coordinates adopts an L1 formulation encoding center coordinates, dimensions, and orientation angles. For lane and road structure analysis, the framework applies analogous supervision with lane classification loss $\mathcal{L}_{mcls}$ and geometric regression loss $\mathcal{L}_{mreg}$, each scaled by task-specific coefficients. The total loss of QFormer is:
\begin{equation}
    \mathcal{L}_{pc} = \lambda_{c}\mathcal{L}_{cls} +\lambda_{r}\mathcal{L}_{reg} + \lambda_{mc}\mathcal{L}_{mcls} + \lambda_{mr}\mathcal{L}_{mreg}, 
\end{equation}
where $\lambda_{c}$, $\lambda_{r}$, $\lambda_{mc}$, and $\lambda_{mr}$ represent loss balancing coefficients for detection classification, detection regression, lane classification, and lane regression tasks, respectively.

For the LLM, we leverage the auto-regressive cross-entropy loss $\mathcal{L}_{ce}$. The unified training objective aggregates these components as:
\begin{equation}
    \mathcal{L}_{total} = \mathcal{L}_{pc} + \mathcal{L}_{ce}.
\end{equation}

\begin{table*}[t!]
    \centering
    \caption{Comparison of open-loop and closed-loop performance on the Bench2Drive dataset. The best results are highlighted in \textbf{bold}. The up arrow ($\uparrow$) indicates that higher is better, while the down arrow ($\downarrow$) indicates that lower is better.}
    \label{tab:main_results}
    \scalebox{0.9}{
    \begin{tabular}{l c c cc c}
        \toprule
        \textbf{Method} & \multicolumn{1}{c}{\textbf{Open-loop Metric}} & \multicolumn{4}{c}{\textbf{Closed-loop Metric}} \\
        \cmidrule(lr){2-2} \cmidrule(lr){3-6}
        & {Avg. L2 $\downarrow$} & {Driving Score $\uparrow$} & {Success Rate(\%) $\uparrow$} & {Efficiency $\uparrow$} & {Comfortness $\uparrow$} \\
        \midrule
        AD-MLP \cite{zhai2023rethinking}  & 3.64 & 18.05 & 0.00  & 48.45  & 22.63 \\
        UniAD-Tiny \cite{hu2023planning} & 0.80 & 40.73 & 13.18 & 123.92 & 47.04 \\
        UniAD-Base \cite{hu2023planning}& 0.73 & 45.81 & 16.36 & 129.21 & 43.58 \\
        VAD \cite{jiang2023vad} & 0.91 & 42.35 & 15.00 & 157.94 & 46.01 \\
        TCP \cite{wu2022trajectory} & 1.70 & 40.70 & 15.00 & 54.26  & 47.80 \\
        ThinkTwice \cite{jia2023think} & 0.95 & 62.44 & 31.23 & 69.33  & 16.22 \\
        DriveAdapter \cite{jia2023driveadapter} & 1.01 & 64.22 & 33.08 & 70.22  & 16.01 \\
        MomAD \cite{song2025don}& 0.87 & 44.54 & 16.71 &\textbf{170.21} &\textbf{48.63}\\
        DriveTransformer \cite{jia2025drivetransformer}& \textbf{0.62} & 63.46 & 35.01 & 100.64 & 20.78 \\
        ReAL-AD \cite{lu2025real}     & 0.84 & 41.17 & 11.36 & {--}   & {--}    \\
        CogAD \cite{wang2025cogad}      & {--} & 48.30 & 24.00 & 142.00 & 40.37 \\
        \midrule
        \textbf{Ours}                   & 0.63 & \textbf{78.21 (22\%$\uparrow$)} & \textbf{56.93 (63\%$\uparrow$)} & 169.52 & 20.50 \\
        \bottomrule
    \end{tabular} }
\end{table*}

\begin{figure*} 
    \centering 

    \begin{subfigure}{0.33\textwidth}
        \centering
        \includegraphics[width=1.0\linewidth]{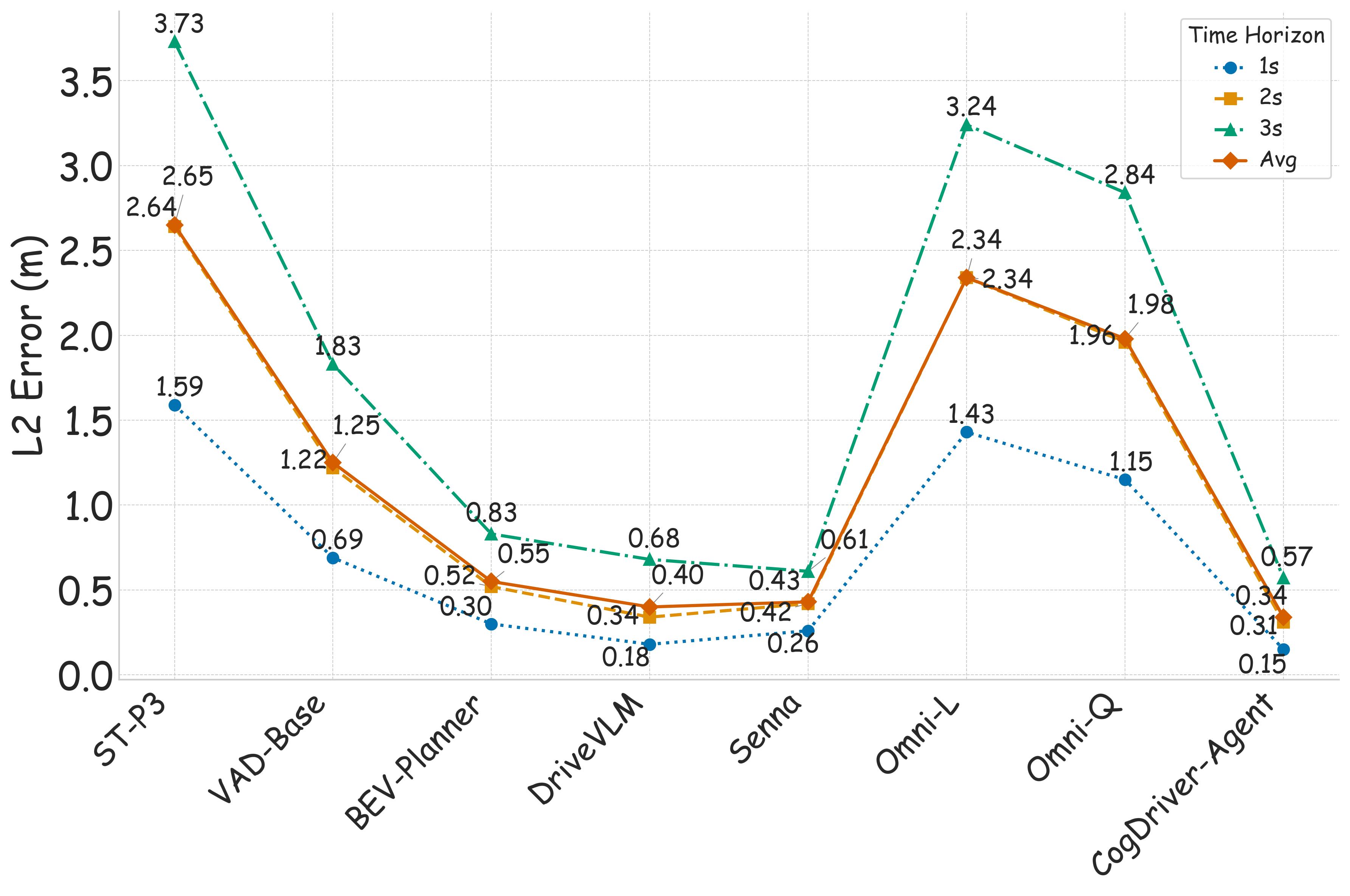}
        \label{fig:l2_error}
    \end{subfigure}
    \begin{subfigure}{0.33\textwidth}
        \centering
        \includegraphics[width=1.0\linewidth]{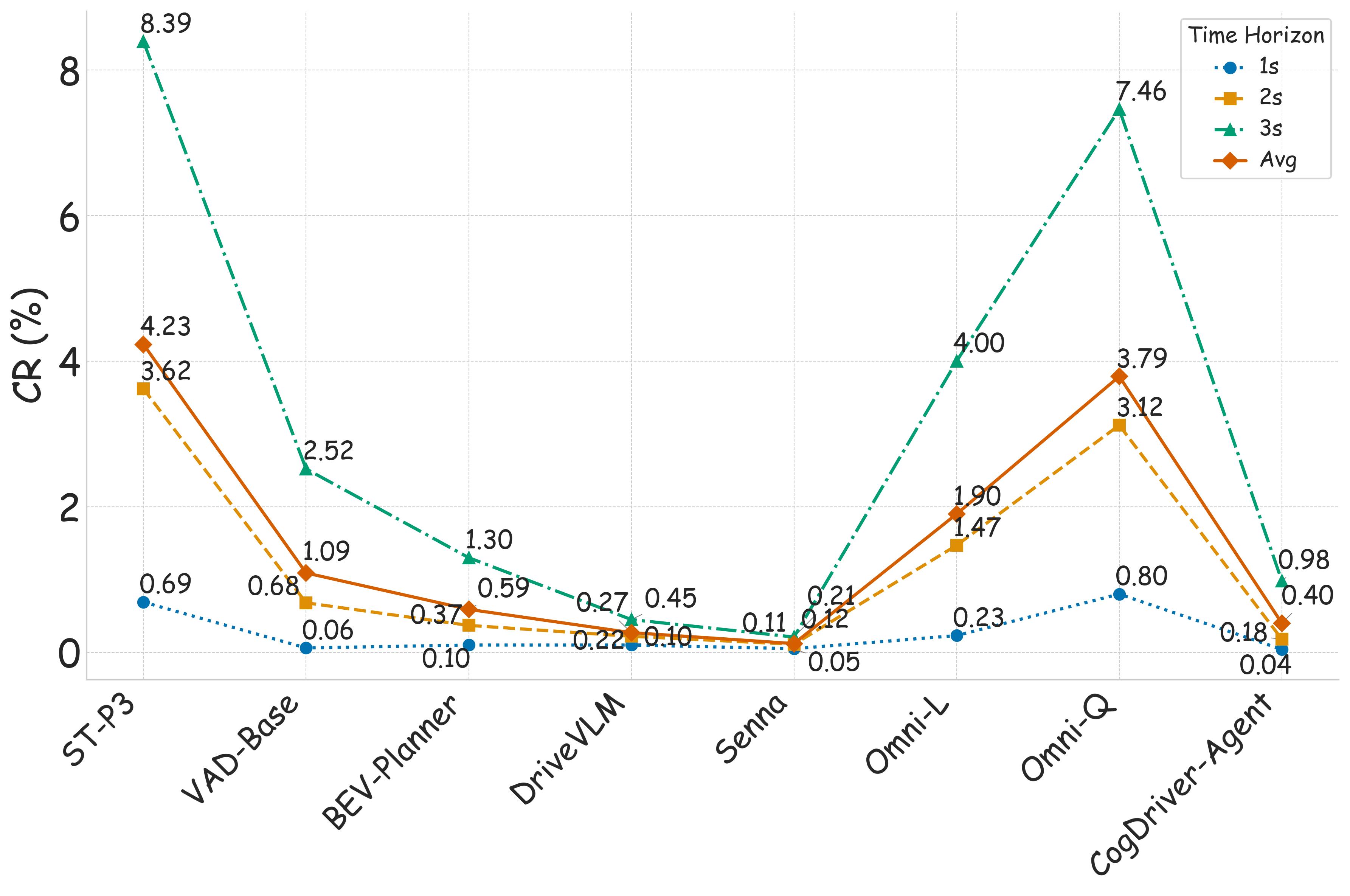}
        \label{fig:cr_rate}
    \end{subfigure}
    \begin{subfigure}{0.33\textwidth}
        \centering
        \includegraphics[width=1.0\linewidth]{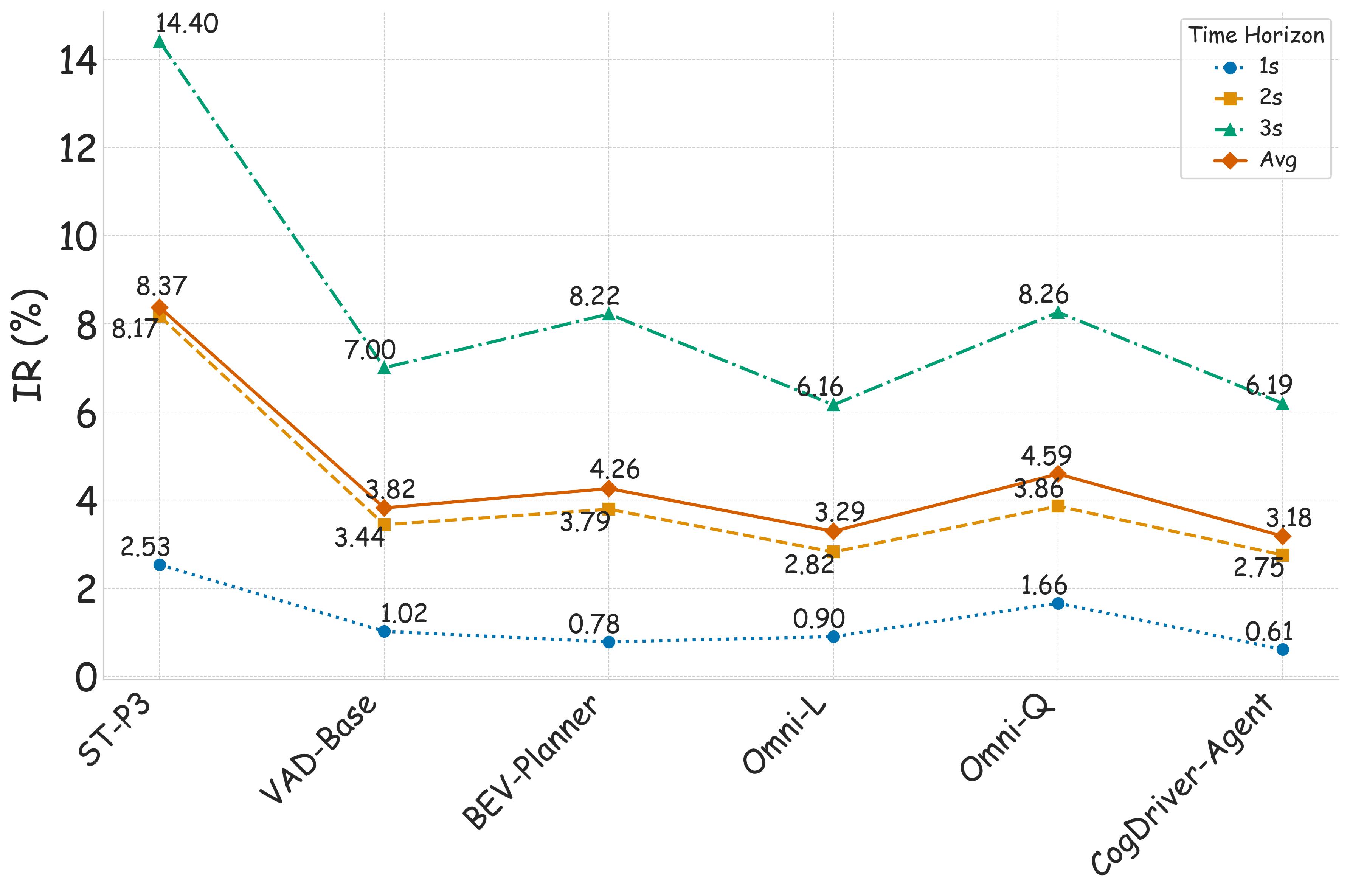}
        \label{fig:ir_rate}
    \end{subfigure}

    \caption{Open-loop planning results on the nuScenes dataset. We compare CogDriver-Agent against several baselines on L2 (left), CR (middle), and IR (right). Performance is reported for 1-3\,s horizons and their average. Lower values are better. Our method demonstrates state-of-the-art or highly competitive performance across all metrics.}
    \label{fig:open}
\end{figure*}

\begin{table*}
    \centering
    \caption{Comparison of VQA performance conducted on the {\model}-nuScenes and {\model}-Bench2Drive datasets. The results show that {\model}-Agent consistently outperforms previous state-of-the-art open-source MLLMs across most metrics. Notably, all reported indicators are positively oriented, with higher values indicating superior performance. CI-r: CIDEr, BL-1:  BLEU-1,  BL-4:  BLEU-4, ME-R: METEOR, RO-L: ROUGE-L.}
    \renewcommand{\arraystretch}{1.1}
    \scalebox{0.9}{
    \begin{tabular}{c|ccccc|ccccc}
    \hline
        \multirow{2}{*}{\textbf{Model}} & \multicolumn{5}{c|}{\textbf{{\model}-nuScenes}} & \multicolumn{5}{c}{\textbf{{\model}-Bench2Drive}} \\
        \cline{2-11}
         ~ &  \textbf{CI-r} & \textbf{BL-1} & \textbf{BL-4} & \textbf{ME-R} & \textbf{RO-L} &  \textbf{CI-r} & \textbf{BL-1} & \textbf{BL-4} & \textbf{ME-R} & \textbf{RO-L}\\
        \hline
        \multicolumn{1}{l|}{Qwen2.5VL 72B \cite{bai2025qwen2}} &67.14  &18.78  &3.25  &20.75  &21.91 &87.57 &28.03 & 5.81 & 67.27 & 28.10\\
        \multicolumn{1}{l|}{Qwen2.5VL 32B} & 59.37 & 15.88& 1.72 & 17.69& 19.13 &83.80 & 30.76 & 5.05 & 58.09 & 25.84\\
        \multicolumn{1}{l|}{Qwen2.5VL 7B} & 62.78& 19.86& 2.96& 22.52& 22.34 & 75.65 &14.55 & 2.20 &71.03 &21.88\\
        \multicolumn{1}{l|}{Qwen2.5VL 3B} & 48.59  & 19.08 &1.91  &22.51 & 21.13 &67.21 &6.39 & 0.92 & 77.27& 19.07\\
        \multicolumn{1}{l|}{Qwen2VL 72B \cite{team2024qwen2}} &  57.10 & 25.76&  4.46& 31.81& 26.56 & 78.60 & 8.21 & 1.67 & \textbf{83.38} & 22.60\\
         \multicolumn{1}{l|}{InternVL3 14B \cite{zhu2025internvl3}}  & 70.01 & 8.82 & 1.09 & 74.15 & 19.18 & 78.55 & 9.78 & 1.71 & 78.42 & 22.2 \\
        \multicolumn{1}{l|}{InternVL3 8B } & 64.64 & 6.41 & 0.67 & 74.02 & 16.7 & 70.58 & 7.33 & 1.1 & 76.56 & 19.42 \\

         \multicolumn{1}{l|}{LLaVA$\_$NEXT 7B \cite{li2024llava}} &53.54 & 4.71 & 0.47 & \textbf{76.09} & 15.63 & 60.08 & 5.35 & 0.63 & 78.53 & 17.51\\
        \hline
        \multicolumn{1}{l|}{{\model}-Agent} & \textbf{92.39} & \textbf{51.54} & \textbf{14.45}& 64.45& \textbf{32.75}& \textbf{95.46} & \textbf{47.25} &\textbf{18.06} &58.78 & \textbf{36.99} \\
        \hline
    \end{tabular}
    }
    
    \label{tab:qa}
\end{table*}

\begin{figure*}
    \centering
    \includegraphics[width=0.9\linewidth]{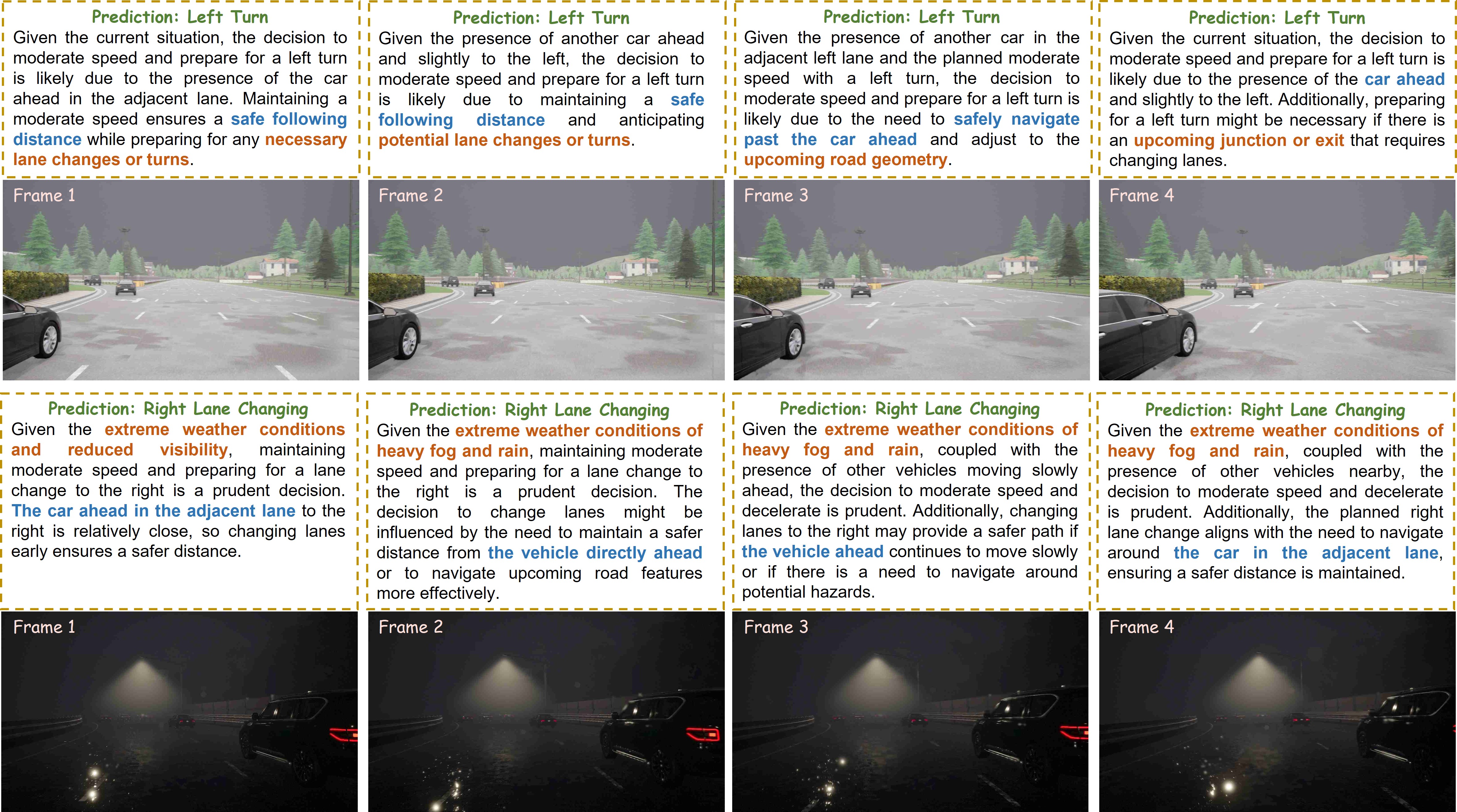}
    \caption{Visualization of Temporally Coherent Reasoning by {\model}-Agent. We present two challenging driving scenarios: a left turn in clear conditions (top) and a right lane change in adverse weather (bottom). For each, we visualize the agent's frame-by-frame narrative predictions. The agent demonstrates cognitive inertia by maintaining a consistent high-level plan (e.g., 'Left Turn'). Crucially, the underlying rationale is not static; it evolves as the scene unfolds, maturing from reacting to a 'car ahead' (Frame 1) to anticipating an 'upcoming junction' (Frame 4), proving its capacity for sophisticated, long-term reasoning.}
    \label{fig:consistency}
\end{figure*}

\section{Experiments}


\subsection{Implementation Details}
For visual feature extraction, we employ EVA-02-L \cite{fang2024eva} as our vision encoder, which is pretrained through masked image modeling with knowledge distillation from CLIP \cite{radford2021learning}. This approach ensures strong alignment between visual features and language representations. LLaVA v1.5 \cite{liu2024improved} serves as our base model, and we adopt the same dataset composition and training configuration for the 2D pretraining phase. During the fine-tuning stage, we optimize the model using AdamW \cite{loshchilov2016sgdr} with a batch size of 16. We apply differentiated learning rates: $4 \times 10^{-4}$ for the projector module, while keeping both the visual encoder and large language model at a lower learning rate of $2\times 10^{-5}$ to preserve their pretrained knowledge. The entire fine-tuning process utilizes a cosine annealing schedule to maintain training stability and achieve optimal convergence. More details are provided in the Appendix.

\begin{table*}[t]
    \centering
    \caption{Ablation study on the \model-nuScenes dataset. We assess \model-Agent's performance with different language components enabled by measuring VQA accuracy and open-loop planning quality, where L2 distance, collision rate, and intersection rate are evaluated as the average values over the 1, 2, and 3\,s horizons.}
    \renewcommand{\arraystretch}{1.2}
    \scalebox{0.9}{
    \begin{tabular}{l|ccccc|ccc|ccc}
    \hline
        \multirow{2}{*}{\textbf{ID}} &  \textbf{Environment} & \textbf{Dynamic} & \textbf{Static} & \multirow{2}{*}{\textbf{Reasoning}} & \multirow{2}{*}{\textbf{Action}} & \multicolumn{3}{c|}{\textbf{VQA}} & \multicolumn{3}{c}{\textbf{Open-Loop}}\\
        \cline{7-12}
        ~ & \textbf{Description} & \textbf{Object} & \textbf{Scene} & ~ &~ & \textbf{BL-1} $\uparrow$ & \textbf{Precision} $\uparrow$ & \textbf{Recall} $\uparrow$ & \textbf{L2} $\downarrow$ & \textbf{CR} $\downarrow$ & \textbf{IR} $\downarrow$ \\
        \hline
        1 &  &$\checkmark$ &$\checkmark$ &$\checkmark$ &$\checkmark$ & 47.91 & 49.00 & 53.01 & 0.34 & \textbf{0.37} & 3.33\\
        2 &  $\checkmark$ & & $\checkmark$ &$\checkmark$ &$\checkmark$  & 50.14 & 54.22 & 52.31 & 0.34 & 0.42 & \textbf{3.08}\\
        3 & $\checkmark$ & $\checkmark$ & &$\checkmark$ &$\checkmark$  & 49.13 & 52.11 & 51.68 & 0.34 & \textbf{0.37} & 3.16\\
        4 & $\checkmark$ &  $\checkmark$ &$\checkmark$ & &$\checkmark$  & 48.25 & 53.43 & 50.95 & 0.34 & 0.43 & 3.22\\
        5 & $\checkmark$ &  $\checkmark$ &$\checkmark$ &$\checkmark$ &  & 50.57 & 52.95 & 53.46 & 0.34 & 0.43 & 3.16\\
        6 & $\checkmark$ &  $\checkmark$ &$\checkmark$ &$\checkmark$ & $\checkmark$ & \textbf{51.54} & \textbf{53.32} & \textbf{55.19} & 0.34 & 0.40 & 3.18\\
        \hline
    \end{tabular}
    }

    \label{tab:ab1}
\end{table*}

\subsection{Closed-Loop Trajectory Planning Task}
As shown in Table~\ref{tab:main_results}, we evaluate our model on the challenging Bench2Drive benchmark for closed-loop trajectory planning. Our method establishes a new state-of-the-art by a significant margin on the primary metrics, achieving a Driving Score of 78.21 and a Success Rate of 56.93\%. These results represent a 22\% and a substantial 63\% relative improvement over the previous best-performing methods, respectively. This dramatic improvement in long-term planning success underscores the effectiveness of instilling cognitive inertia, as our agent maintains coherent intentions and avoids the decision jitter that plagues prior works. Furthermore, our model remains highly competitive on other key metrics, achieving the second-best open-loop Avg. L2 score (0.63 vs. 0.62) and a comparable Efficiency score (169.52 vs. 170.21), demonstrating its all-around strong performance without sacrificing imitation quality or planning efficiency.

\subsection{Open-Loop Trajectory Planning Task}

As shown in Fig. \ref{fig:open}, we compare our proposed {\model}-Agent against a comprehensive set of state-of-the-art open-loop planning baselines on the nuScenes dataset, including ST-P3 \cite{hu2022st}, VAD \cite{jiang2023vad}, BEV-Planner \cite{li2024ego}, DriveVLM \cite{tian2024drivevlm}, Senna \cite{jiang2024senna}, and OmniDrive (Omni-L, Omni-Q) \cite{wang2025omnidrive}. The proposed {\model}-Agent achieves the lowest average L2 distance of 0.34\,m, outperforming other baselines such as BEV-Planner with 0.55\,m, DriveVLM with 0.40\,m, and VAD-Base with 1.25\,m. In terms of collision rate (CR), {\model}-Agent attains an average value of 0.40\%, which is slightly higher than Senna's best result of 0.12\% and DriveVLM's 0.27\%, but remains lower than most other baselines. For infraction rate (IR), {\model}-Agent sets a new state-of-the-art with an average of 3.18\%, outperforming all previous methods reporting this metric, including BEV-Planner with 4.26\%, VAD-Base with 3.82\%, and Omni-Q with 4.59\%. Across all evaluated horizons, {\model}-Agent demonstrates superior or highly competitive performance in both safety and trajectory accuracy, highlighting the effectiveness of its reasoning framework under fair evaluation protocols that prevent explicit ego status leakage.

\subsection{Driving Visual Question Answering Task}
As shown in Table \ref{tab:qa}, {\model}-Agent achieves significant performance improvements across all metrics on both {\model}-nuScenes and {\model}-Bench2Drive benchmarks compared to state-of-the-art open-source MLLMs. On {\model}-nuScenes, our method demonstrates particularly strong results with a 37.6\% relative improvement in CIDEr, while showing even more dramatic gains of 100.1\% and 224.0\% in BLEU-1 and BLEU-4 scores, respectively. The performance on {\model}-Bench2Drive remains consistently superior, with {\model}-Agent achieving 9.0\%, 53.6\%, and 210.8\% relative improvements in CIDEr, BLEU-1, and BLEU-4 metrics over the best baseline. Notably, these substantial advances are consistently reflected across both datasets in the ROUGE-L metric as well. These comprehensive results not only validate {\model}-Agent's enhanced multimodal reasoning capabilities but also establish new state-of-the-art performance for VQA tasks in autonomous driving scenarios.

\subsection{Ablation Study}

\subsubsection{Analysis on Different Language Components.}
Table \ref{tab:ab1} ablates the language components of {\model}-Agent. The full configuration yields the best performance with a 51.54 BLEU-1 score. Individual components provide targeted benefits: environment context boosts language generation by 7.6\% in BLEU-1, while dynamic and static object descriptions are critical for safety, achieving the lowest collision rate of 0.37 and intersection rate of 3.08, respectively. Importantly, trajectory prediction remains robust across all configurations with a stable 0.34 L2 error, demonstrating that enhanced reasoning does not compromise planning accuracy.

\subsubsection{Effectiveness on Temporal Memory Module.}

Table~\ref{tab:tmp} presents the results of the ablation study on the sparse temporal memory module. Introducing the temporal memory module leads to improvements in open-loop driving, as shown by decreases in L2 distance, collision rate, and intersection rate when compared to the baseline without memory. Specifically, L2 distance is reduced from 0.38 to 0.34, collision rate decreases from 0.44 to 0.40, and intersection rate drops from 3.65 to 3.18. While there is a slight reduction in VQA BL-1 and precision, the recall improves, indicating that the model becomes more sensitive to relevant temporal information. These results demonstrate that the temporal memory module enhances sequential reasoning and contributes to more accurate and safer driving performance.

\begin{table}[]
    \centering
    \caption{Ablation study on sparse temporal memory module. L2 distance, collision rate, and intersection rate are evaluated as the average values over the 1, 2, and 3\,s horizons.}
    \renewcommand{\arraystretch}{1.2}
    \scalebox{0.8}{
    \begin{tabular}{c|ccc |ccc}
    \hline
    \multirow{2}{*}{\textbf{TCM}}  & \multicolumn{3}{|c}{\textbf{VQA}} & \multicolumn{3}{|c}{\textbf{Open-Loop}} \\
    \cline{2-7}
    ~& \textbf{BL-1} $\uparrow$& \textbf{Precision} $\uparrow$& \textbf{Recall} $\uparrow$& \textbf{L2} $\downarrow$& \textbf{CR} $\downarrow$& \textbf{IR} $\downarrow$\\
    \hline
        & 52.24 & 54.84 & 54.86 & 0.38 & 0.44 & 3.65\\
    \checkmark & 51.54 & 53.32 & 55.19 & 0.34 & 0.40 & 3.18 \\
    \hline
    \end{tabular}
    }
    
    \label{tab:tmp}
\end{table}

\subsubsection{Inference Efficiency and Throughput}
We evaluate the inference efficiency of {\model}-Agent on a single NVIDIA A800 GPU. As shown in Table \ref{tab:real-time}, compared to the Qwen2.5VL 32B baseline, {\model}-Agent achieves significantly higher throughput, reaching 3410.81 input tokens/s and 391.43 output tokens/s. This high processing speed confirms that our architecture satisfies the strict latency constraints required for real-time autonomous driving.

\begin{table}[h]
    \centering
     \caption{Speed comparison of VLMs.}
     \scalebox{0.8}{
    \begin{tabular}{c|cc}
    \toprule
         \multirow{2.6}{*}{Model} & \multicolumn{2}{c}{pixels = 256 $\times$ 256, tokens = 300} \\
         \cmidrule{2-3}
         ~ & Speed input (tokens/s)& Speed output (tokens/s)\\
         \midrule
         
         Qwen2.5VL 32B & 970.94	& 168.23 \\
         \midrule
         CogDriver-Agent	&\textbf{3410.81}	&\textbf{391.43} \\
    \bottomrule
    \end{tabular}
    \label{tab:real-time}}
\end{table}

\subsection{Qualitative Results}
Fig. \ref{fig:consistency} provides a qualitative analysis of our agent's reasoning process, demonstrating both decision coherence and causal cognitive coherence.

In the 'Left Turn' scenario (top), the agent exhibits strong decision coherence by consistently maintaining the high-level plan 'Left Turn' across all frames, avoiding the decision jitter common in reactive models. More critically, the figure reveals the agent's causal cognitive coherence: the underlying rationale is not static but evolves as the scene unfolds. The reasoning matures from a reactive concern for the 'car ahead' (Frame 1) to a strategic, forward-looking plan based on the 'upcoming junction' (Frame 4). This demonstrates a crucial capability: the agent's internal 'story' of why it is acting becomes more sophisticated as it gathers more evidence. A similar evolution from immediate hazard avoidance to strategic path planning is observed in the adverse weather scenario (bottom). This stands in stark contrast to stateless models, whose reasoning would remain fixated on the most immediate visual cues without forming a coherent, evolving causal narrative. Additional results are provided in the Appendix. 

\section{Conclusion}
This work presents {\model}, a comprehensive framework for advancing interpretable and reliable autonomous driving. By introducing large-scale VLA datasets with rich, temporally dense language annotations and proposing the {\model}-Agent model that distills human-like priors and causal reasoning, we enable more context-aware and explainable driving decisions. Extensive experiments demonstrate that our approach significantly improves performance in safety, comfort, and explanation quality, bringing autonomous vehicles closer to human-level understanding. 

\clearpage
{
    \small
    \bibliographystyle{ieeenat_fullname}
    \bibliography{main}
}

\clearpage
\setcounter{page}{1}
\maketitlesupplementary

\section{{\model}-Data}\label{app:dataset}

\begin{figure*}[tbp]
    \centering

    \begin{subfigure}[b]{0.9\textwidth}
        \centering

        \includegraphics[width=1.0\textwidth, height=0.8\textheight, keepaspectratio]{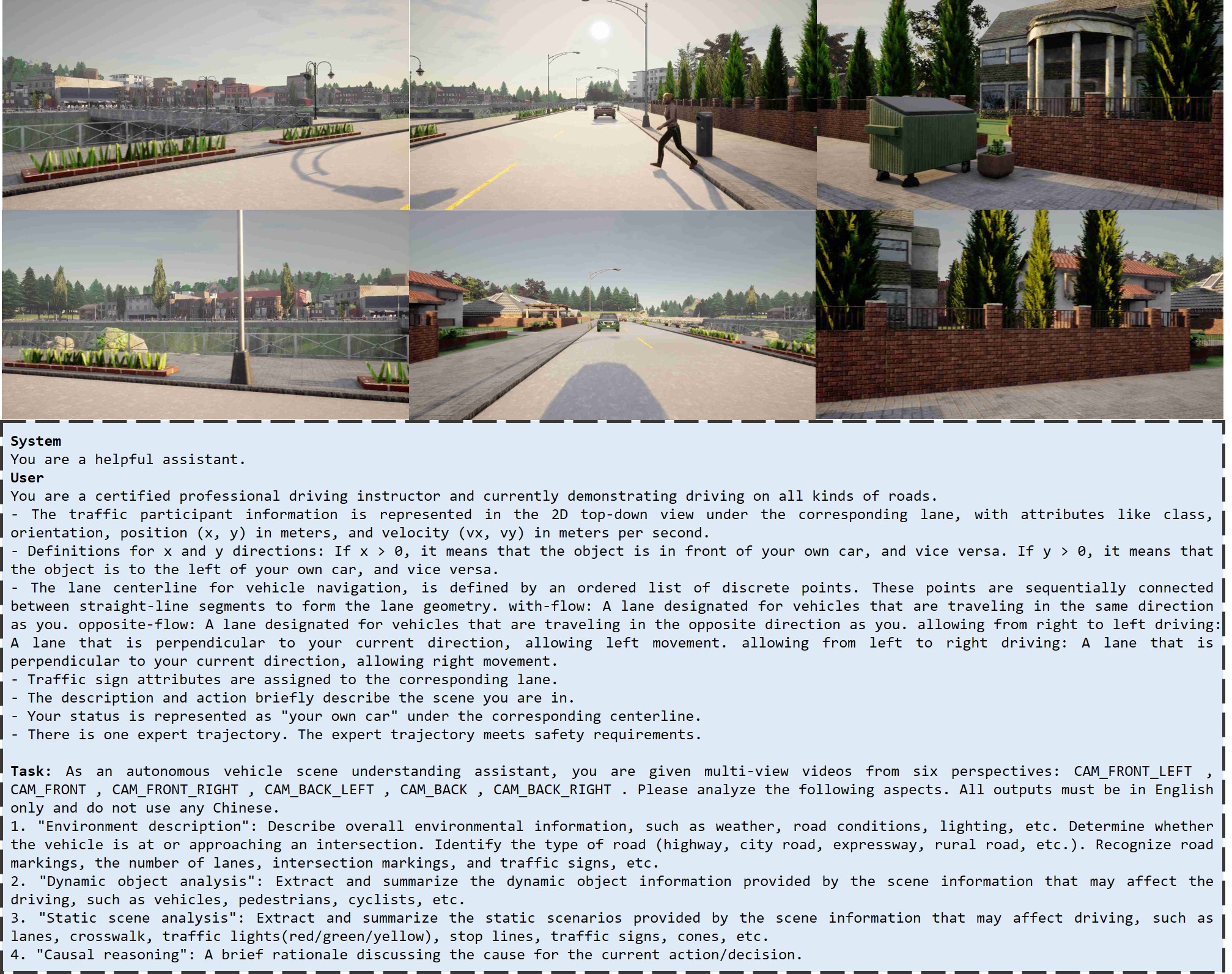}
        \caption{Example of input and system prompt. }
        \label{fig:sub1} 
    \end{subfigure}

    \vspace{1em} 

    \begin{subfigure}[b]{0.9\textwidth}
        \centering
        \includegraphics[width=1.0\textwidth, height=0.8\textheight, keepaspectratio]{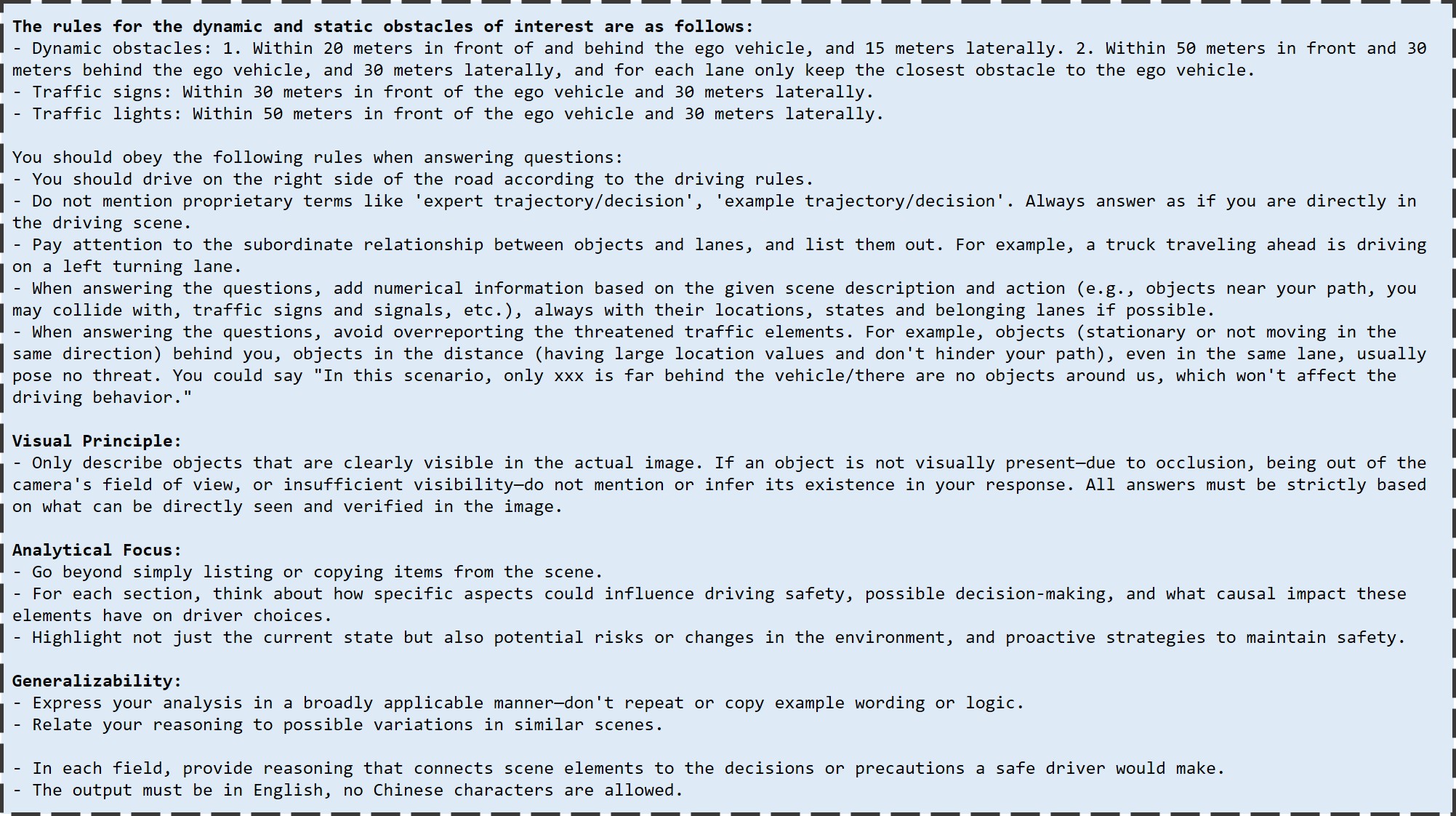}
        \caption{Example of human-centric rules.}
        \label{fig:sub2}
    \end{subfigure}    
\end{figure*}

\begin{figure*}[tbp] 
    \ContinuedFloat 
    \centering

    \begin{subfigure}[b]{0.9\textwidth}
        \centering
        \includegraphics[width=1.0\textwidth, height=0.4\textheight, keepaspectratio]{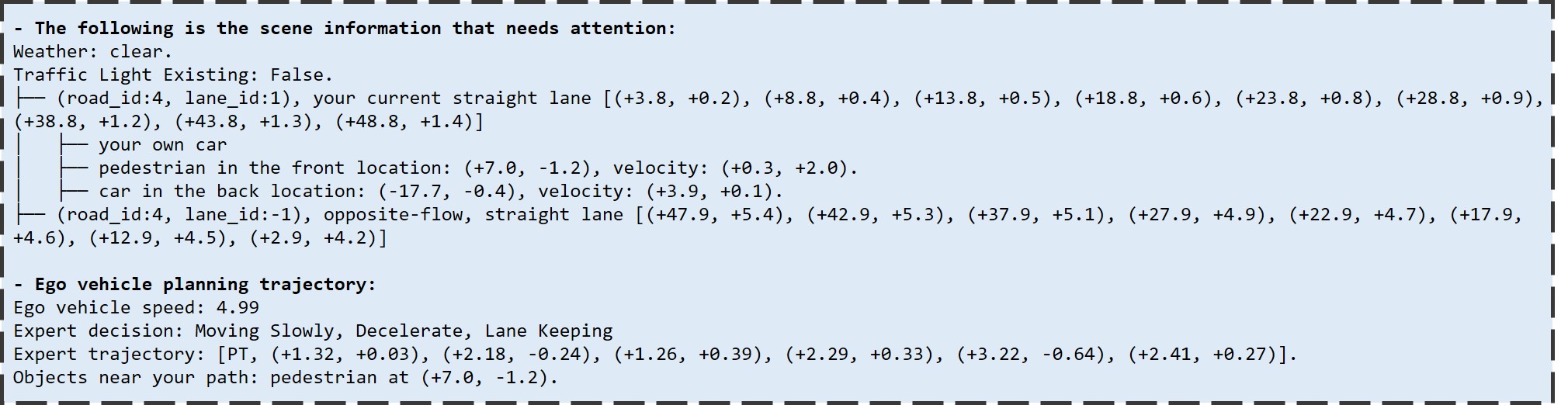}
        \caption{Example of attention and ego states.} 
        \label{fig:sub3}
    \end{subfigure}
    \caption[]{Prompt template and input structure for data annotation. }
    \label{fig:prompt-template} 
\end{figure*}

\subsection{Human-Centric Reasoning Priors}
To further enhance the interpretability and human-likeness of the generated captions, we introduce human-centric reasoning priors. This component incorporates explicit prompts centered around driving rules, visual reasoning principles, and scenario-specific heuristics, as shown in Fig.~\ref{fig:sub2}. By embedding these knowledge-driven cues, the captioning process is steered toward producing language and rationales that reflect the thought process and intuition of an experienced human driver. This augmentation allows the dataset to capture high-level explanations, intention recognition, and structured abstraction that go beyond the descriptive capability of rule-based templates alone, resulting in more instructive and contextually aware supervision for VLA research.

\subsection{Scene-Aware Grounding}
The foundation of our annotation pipeline is Scene-Aware Grounding, a process designed to combat model hallucination by providing an objective, factual basis for all subsequent reasoning. To achieve this, we systematically extract verifiable data directly from the driving logs, rather than relying on a model to infer basic facts. As illustrated in Fig.~\ref{fig:sub3}, this includes precise map context, vehicle dynamics, trajectory profiles, and the states of surrounding agents and traffic signals. This objective information is then structured into a dedicated component of our prompt, forcing the MLLM to condition its output on this unimpeachable ground truth. This methodology ensures that all generated narratives are firmly rooted in the physical reality of the scene, drastically reducing factual errors and providing a reliable foundation for VLA research.

\subsection{Narrative Generation with Cognitive Inertia} 
The generation of narratives with cognitive inertia hinges on the model's ability to ground static principles in dynamic, unfolding events. To enable this, we condition our powerful MVST-MLLM (72B) simultaneously on the structured prompts from our scene-aware grounding and expert driving principles, and the multi-view, multi-frame video itself, as highlighted in Fig.~\ref{fig:prompt-template}.

Crucially, unlike single-frame captioning solutions that can only produce reactive descriptions, our method operates on temporal windows (5 frames). This is the key mechanism that allows the model to perceive not just a state, but a state transition. It learns to connect the "before" and "after," enabling it to generate explanations grounded in causality and persistent intent, the very essence of cognitive inertia. For each 30-second driving scene, overlapping temporal windows comprehensively sample the spatiotemporal dynamics. The resulting output is therefore not a mere caption, but a narrative fragment that explains the current action by linking it to past observations and future goals. Leveraging large-scale distributed inference, our pipeline efficiently produces over 227k such temporally grounded and causally enriched narratives daily, establishing a new standard for high-quality VLA resources.

\subsection{Data Analysis}

\subsubsection{Environment Analysis.}

\begin{figure*}
    \centering
    \includegraphics[height=0.3\textwidth, keepaspectratio]{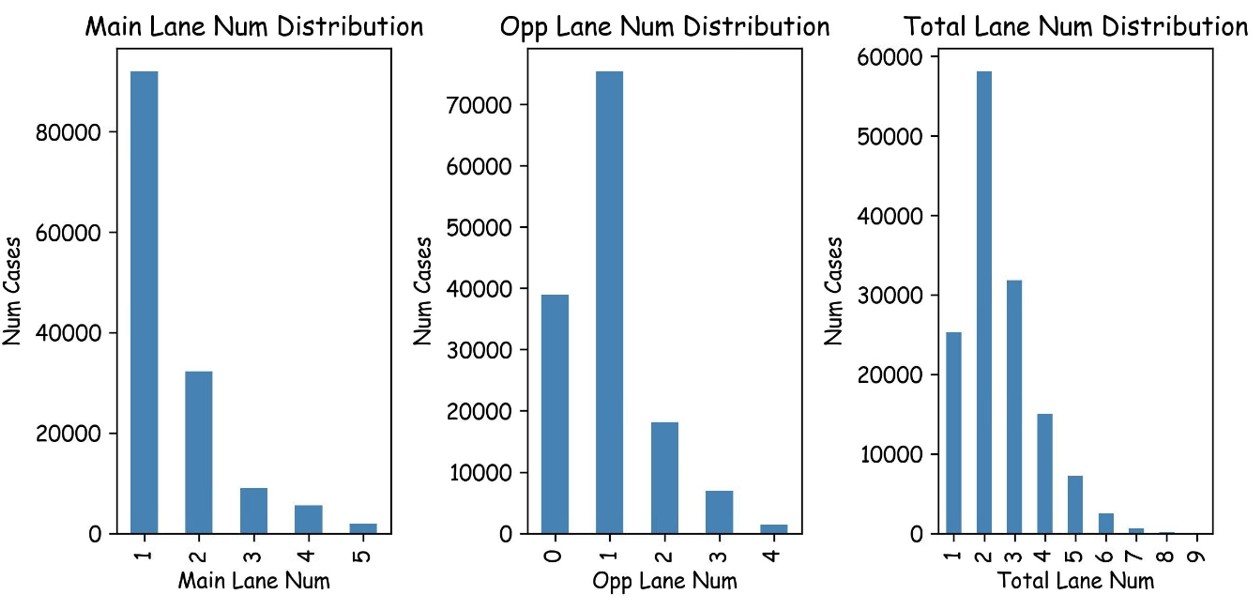}
    \caption{Lane number distributions in the Bench2Drive dataset. The left plot shows the distribution of main (ego-moving direction) lane numbers, with the majority of cases having one or two lanes in the same direction. The middle plot displays the distribution of opposite-direction lanes, where most scenarios involve a single or no opposite lane. The right plot shows the total number of lanes, indicating that most roads in the dataset have two to three lanes in total.}
    \label{fig:lane}
\end{figure*}

\begin{figure*}
    \centering
    \includegraphics[width=0.8\linewidth]{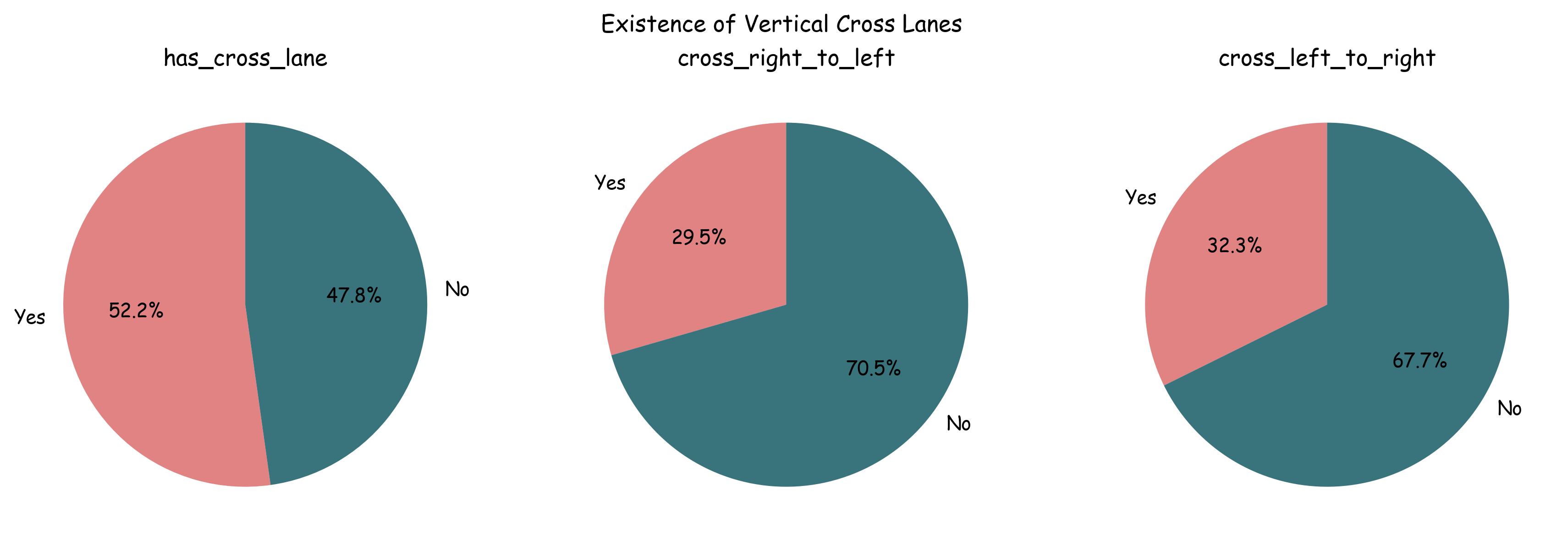}
    \caption{Distribution of cross lanes in the Bench2Drive dataset. The left pie chart shows the proportion of samples with cross lanes (has$\_$cross$\_$lane). The middle and right pie charts illustrate the existence of vertical cross lanes from right to left (cross$\_$right$\_$to$\_$left) and from left to right (cross$\_$left$\_$to$\_$right).}
    \label{fig:cross_lane}
\end{figure*}

\begin{figure*}
    \centering
    \includegraphics[width=0.8\linewidth]{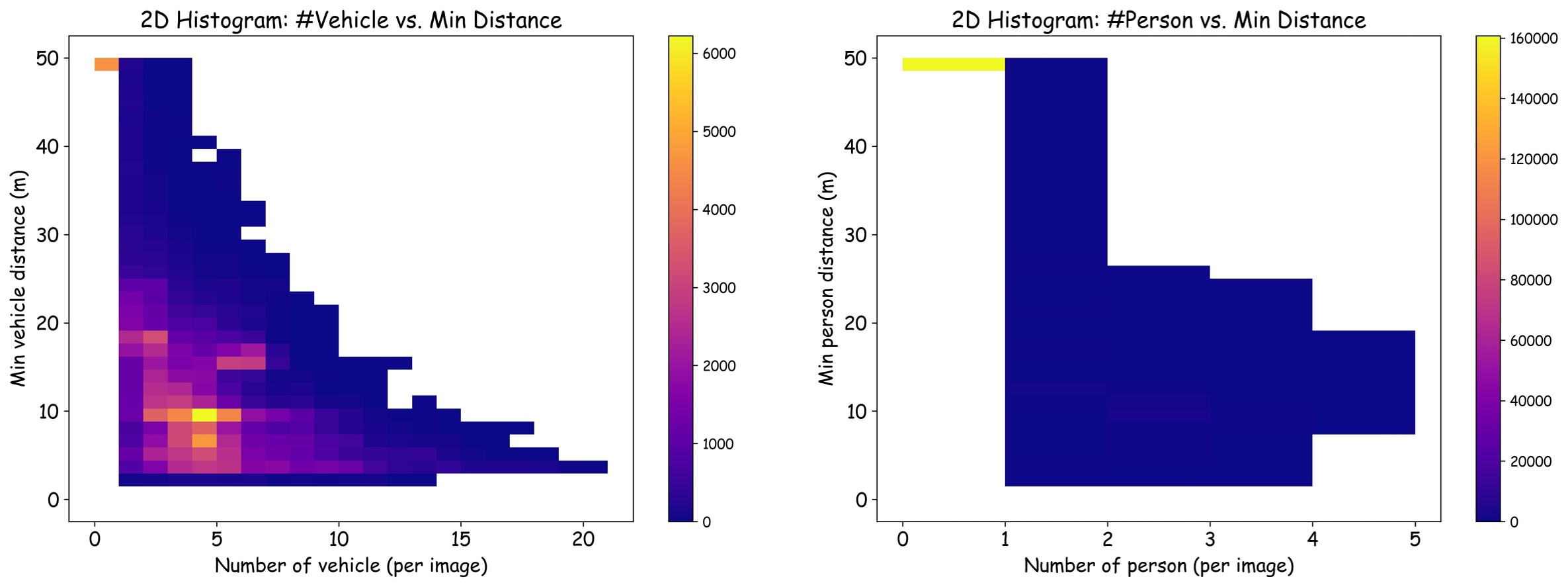}    
    \caption{2D histograms visualizing the relationship between the number of objects and their minimum distance to the ego vehicle in each image. The left plot shows the distribution for vehicles, indicating that as the number of vehicles in the scene increases, the minimum distance to the nearest vehicle generally decreases. The right plot shows the distribution for pedestrians, where most images contain zero or one person, and the minimum distance to the nearest person is typically large. The color bars represent the number of cases in each bin.}
    \label{fig:num}
\end{figure*}

Due to the large data volume and high diversity of Bench2Drive, we primarily showcase analysis on the Bench2Drive dataset.
Fig.~\ref{fig:lane} and Fig.~\ref{fig:cross_lane} illustrate the structural diversity and complexity of road layouts within Bench2Drive. As shown in Fig.~\ref{fig:lane}, the dataset encompasses a wide range of lane configurations, with most scenes featuring one or two same-direction lanes and varying numbers of opposite-direction lanes, leading to total lane counts that are predominantly two or three, but also covering higher numbers. This reflects a substantial spectrum of real-world driving environments. Fig.~\ref{fig:cross_lane} further investigates the occurrence of cross lanes, revealing that approximately half of the samples contain intersecting lanes, thereby capturing both regular and complex intersection scenarios. The breakdown of cross lane types, including left-to-right and right-to-left vertical crossings, highlights the presence of multifaceted intersection designs. Collectively, these characteristics demonstrate that Bench2Drive offers rich and challenging road structures, supporting comprehensive evaluation of autonomous driving models across a diverse range of traffic scenarios. Fig.~\ref{fig:num} presents the joint distribution of the number of surrounding vehicles and pedestrians per frame alongside their minimum distance to the ego vehicle, offering a comprehensive view of both scene density and interaction complexity. The left subplot demonstrates that frames with a larger number of vehicles are associated with reduced minimum distances, reflecting a diverse spectrum of dense and potentially challenging traffic scenarios where close-proximity interactions are common. In contrast, the right subplot shows that while most frames feature zero or one pedestrian, there is still a notable range of cases with multiple pedestrians at varying distances, capturing complex urban scenes where pedestrian interactions can occur. Overall, these results highlight the dataset's rich diversity, spanning from dense multi-vehicle interactions to sparse pedestrian scenarios, while also capturing the realistic distribution of safety-critical events, which are infrequent and typically occur at greater distances from the ego vehicle.

\subsubsection{Action Analysis.} 
\begin{figure*}
    \centering
    \begin{subfigure}[b]{0.49\textwidth}
        \centering
        \includegraphics[width=\linewidth]{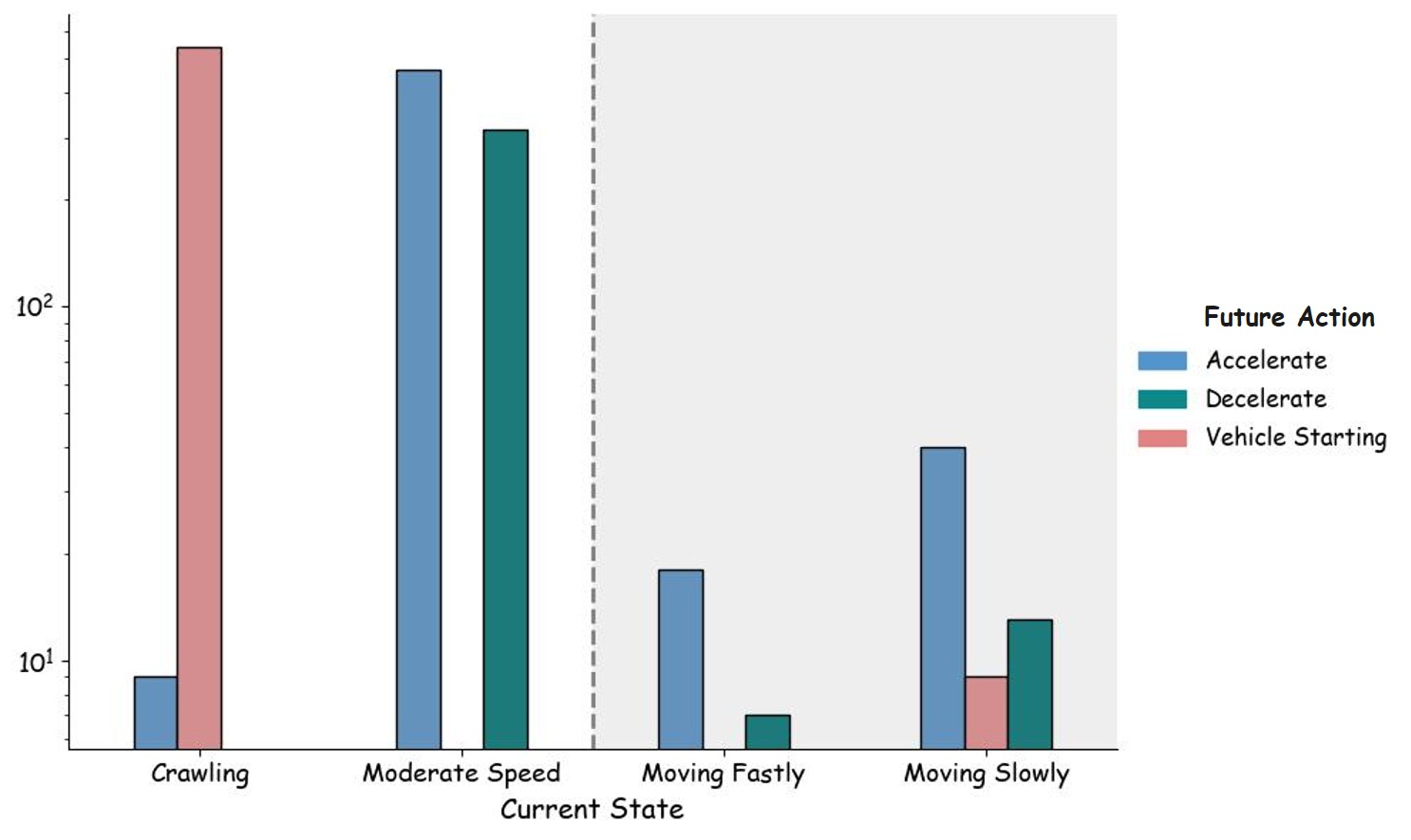}
    \end{subfigure}
    \hfill
    \begin{subfigure}[b]{0.49\textwidth}
        \centering
        \includegraphics[width=\linewidth]{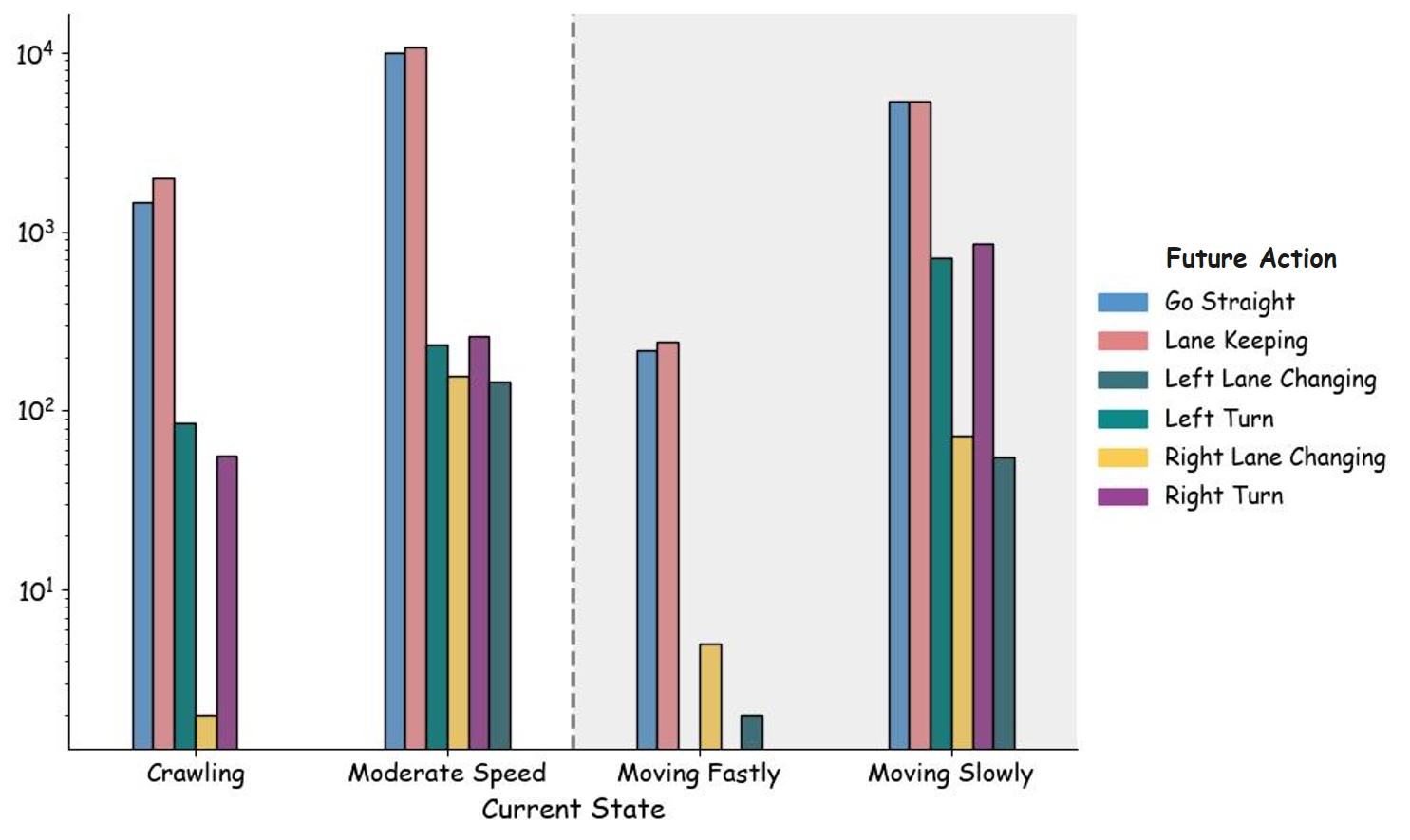}
    \end{subfigure}
    \caption{Ego vehicle's future action at different current states on the nuScenes dataset.}
    \label{fig:data-action-nu}
\end{figure*}

\begin{figure*}
    \centering
    \begin{subfigure}[b]{0.49\textwidth}
        \centering
        \includegraphics[width=\linewidth]{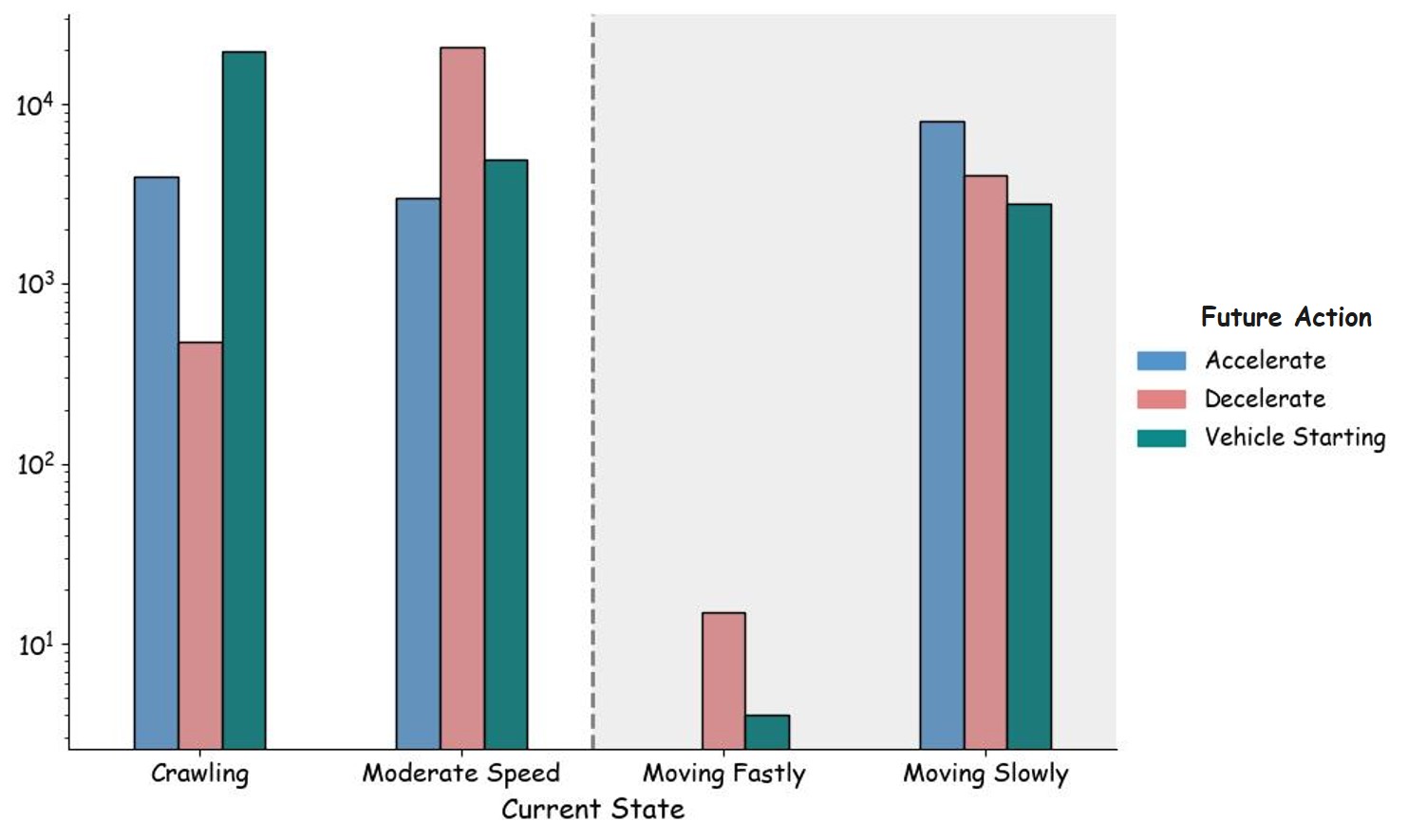}
    \end{subfigure}
    \hfill
    \begin{subfigure}[b]{0.49\textwidth}
        \centering
        \includegraphics[width=\linewidth]{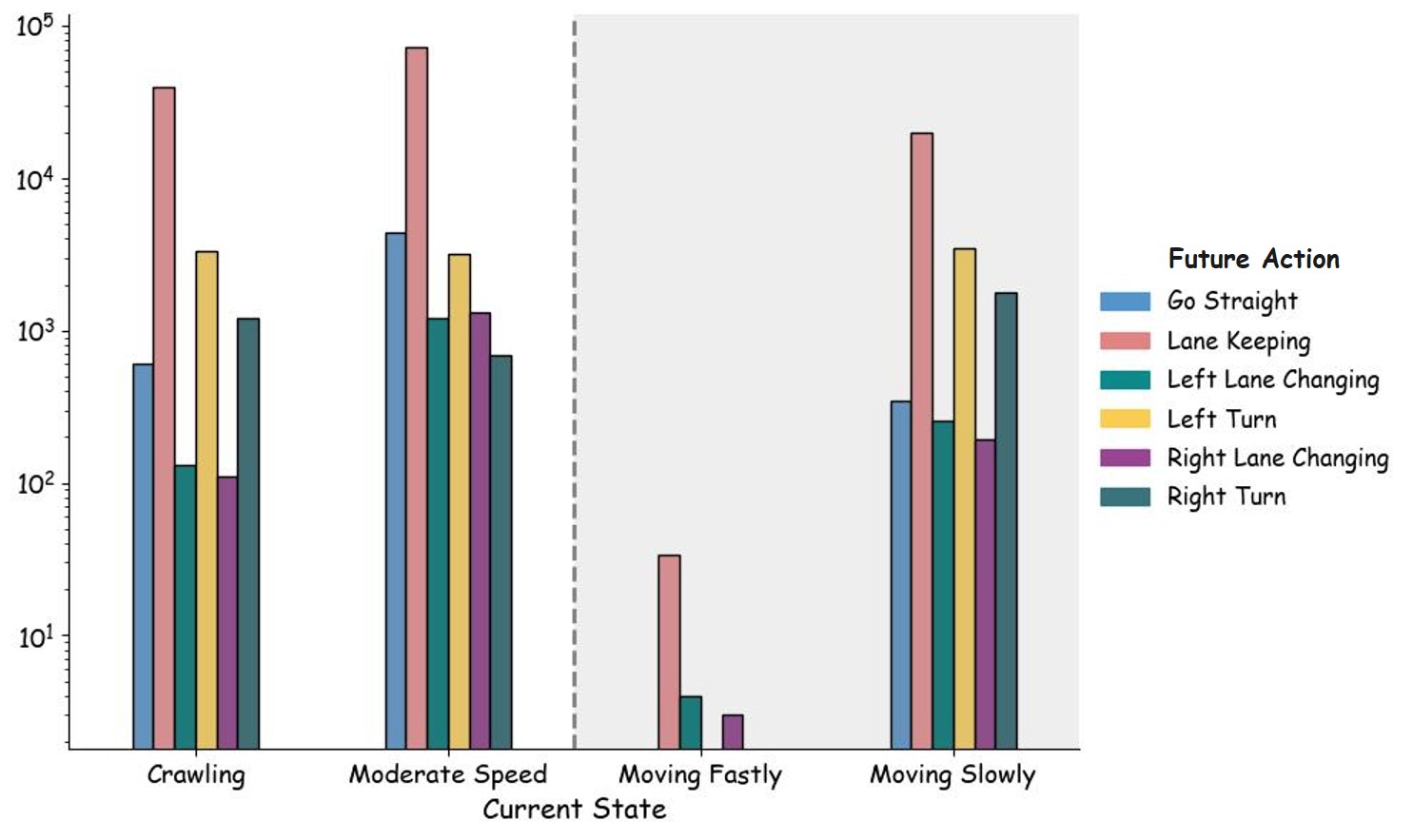}
    \end{subfigure}
    \caption{Ego vehicle's future action at different current states on the Bench2Drive dataset.}
    \label{fig:data-action-b2d}
\end{figure*}

Fig.~\ref{fig:data-action-nu} and Fig.~\ref{fig:data-action-b2d} illustrate the distribution of ego vehicle future actions along two key dimensions: the high-level longitudinal action category (such as accelerating, decelerating, or vehicle starting) and the fine-grained maneuver type (such as lane keeping, lane changing, and turning), conditioned on the current driving state.

Across both the nuScenes and Bench2Drive datasets, similar trends are observed. When the ego vehicle is in the Crawling state, Vehicle Starting is the most common future action, reflecting typical behavior as vehicles transition from a stop. As the current state shifts to Moderate Speed, future actions become more evenly distributed, with Accelerate and Decelerate gaining prominence, indicative of dynamic adjustments in flowing traffic. In the Moving Fast state, the repertoire of future actions broadens, yet the incidence of Vehicle Starting diminishes significantly in both datasets.

The right panels of each figure provide further granularity, showing distributions of specific maneuvers. Here, Go Straight overwhelmingly dominates at moderate and high speeds, while more complex behaviors such as lane changes and turns are less frequent. Notably, the Bench2Drive dataset exhibits a higher proportion of lane change and turning maneuvers, particularly in low-speed states, suggesting greater environmental diversity and more elaborate behavior annotations compared to nuScenes. Overall, these results demonstrate both consistent trends and meaningful differences across datasets, highlighting the diversity of scene contexts and the complexity of decision-making captured within each.

\subsubsection{Causal Reasoning Analysis}
\begin{figure*}
    \centering
   \includegraphics[width=0.8\linewidth]{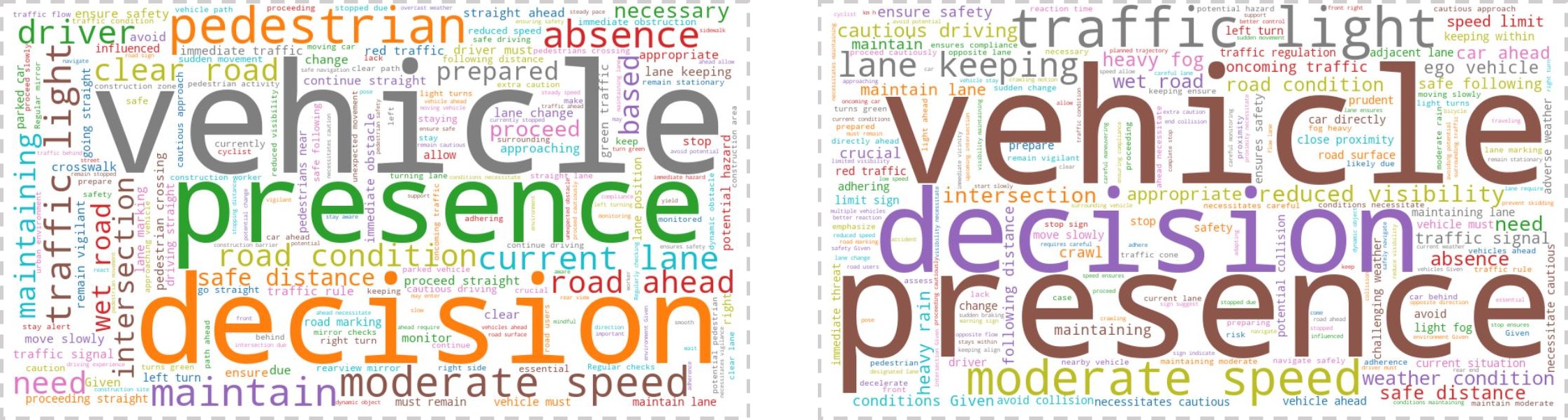}
    \caption{Word cloud of causal reasoning annotations on the nuScenes (left) and Bench2Drive (right) dataset. }
    \label{fig:reason}
\end{figure*}

Fig.~\ref{fig:reason} presents word clouds of causal reasoning annotations for the nuScenes (left) and Bench2Drive (right) datasets, highlighting the core concepts and decision factors that guide autonomous driving behavior in each setting. Prominent terms such as "vehicle," "presence," and "decision" appear in both datasets, reflecting a shared emphasis on situational awareness and decision-making. Additional keywords like "pedestrian," "traffic light," "moderate speed," and "safe distance" reveal considerations of dynamic agents, traffic regulations, and safety constraints. The distribution and diversity of these terms illustrate the contextual richness and the range of challenges addressed by causal reasoning in complex real-world scenarios across both datasets.

\subsubsection{Case Study.}

\begin{figure*}[tbp]
    \centering
    \includegraphics[width=1.0\linewidth]{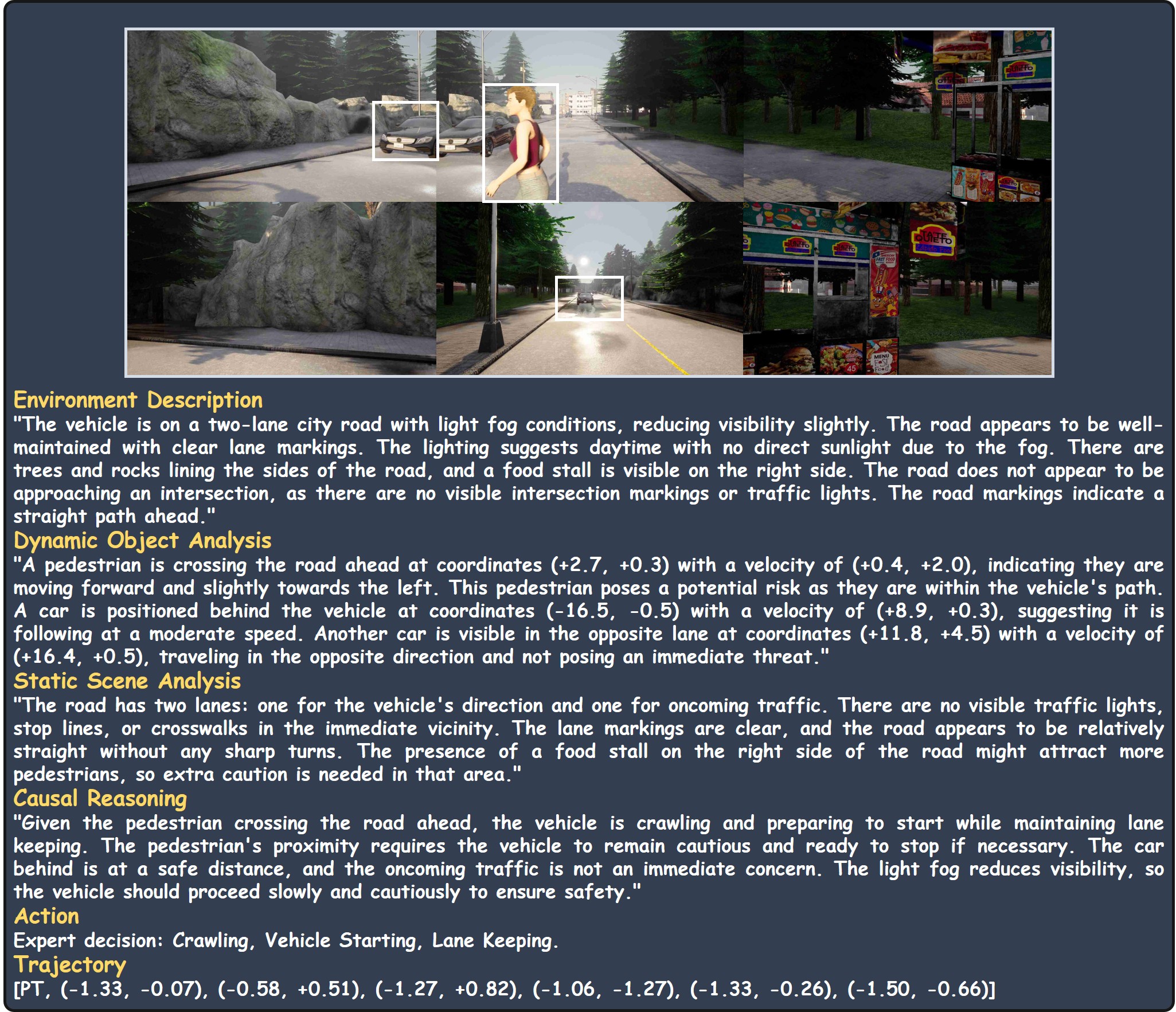}
    \caption{A dynamic object crossing example from the {\model}-Bench2Drive dataset is illustrated. The white box represents the dynamic objects around the ego vehicle.}
    \label{fig:case-b2d}
\end{figure*}

\begin{figure*}[tbp]    
\centering
    \includegraphics[width=1.0\linewidth]{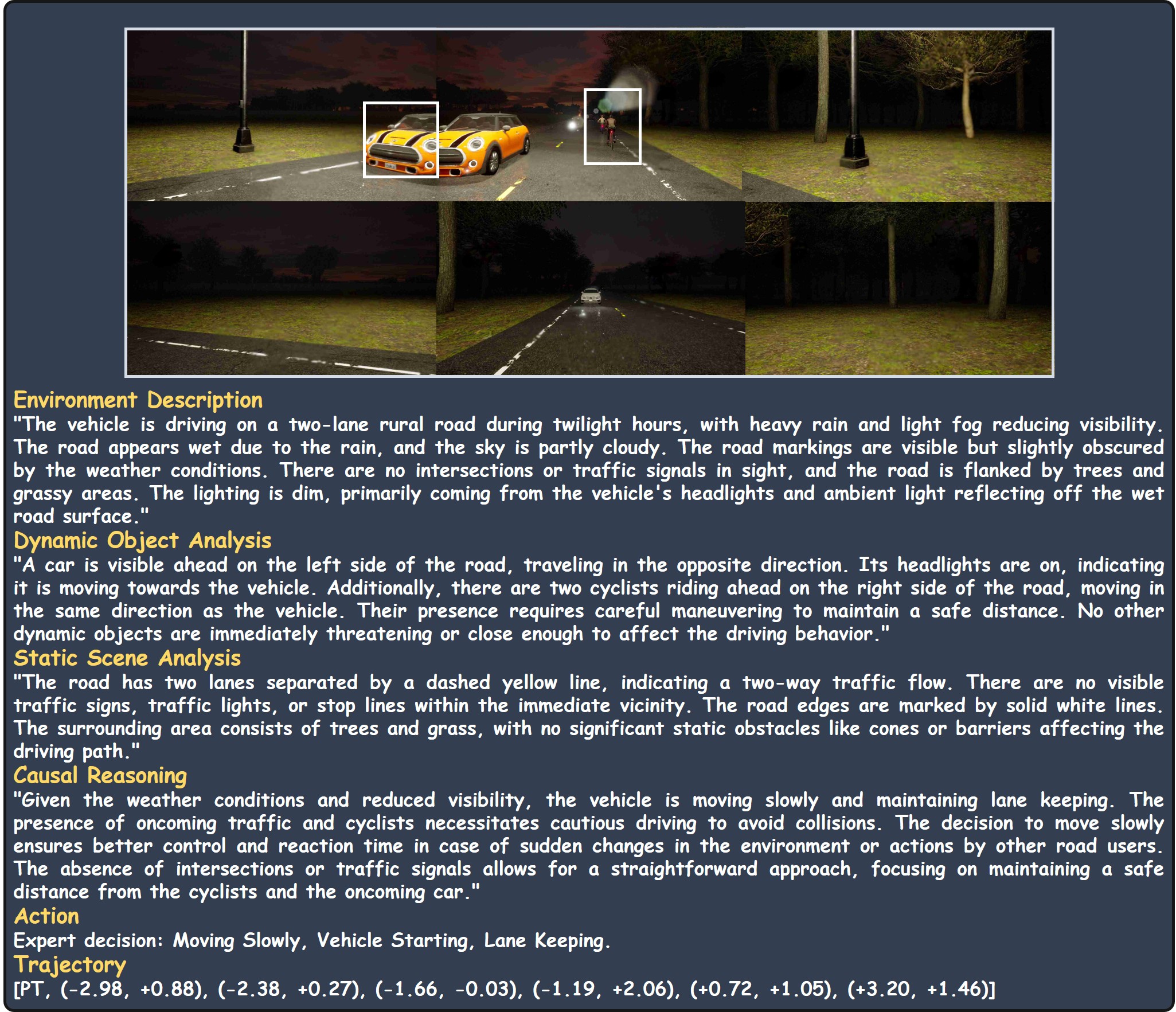}
    \caption{A hazard at side lane two ways example from the {\model}-Bench2Drive dataset is illustrated. The white box represents the dynamic objects around the ego vehicle.}
    \label{fig:case1-b2d}
\end{figure*}

Fig.~\ref{fig:case-b2d} and Fig.~\ref{fig:case1-b2d} from the {\model}-Bench2Drive dataset showcase the annotation model’s scene understanding capability and ability to reason effectively in complex and challenging driving scenarios. Fig.~\ref{fig:case-b2d} depicts a scenario where the ego vehicle navigates a foggy two-lane city road as a pedestrian crosses its path, requiring the vehicle to proceed slowly and remain ready to stop. Additional dynamic agents, such as a following vehicle and an oncoming car, further complicate the scene. Fig.~\ref{fig:case1-b2d} presents a rural road during twilight with wet conditions, reduced visibility, and multiple hazards, including oncoming traffic and cyclists. In both cases, the lack of traffic signals and the presence of dynamic obstacles highlight the importance of cautious, context-aware planning. These examples demonstrate the dataset’s capacity to capture realistic challenges and support the development of interpretable, safety-focused autonomous driving models.

\clearpage
\begin{figure*}   
\centering
    \includegraphics[width=0.85\linewidth, height=1.25\textwidth]{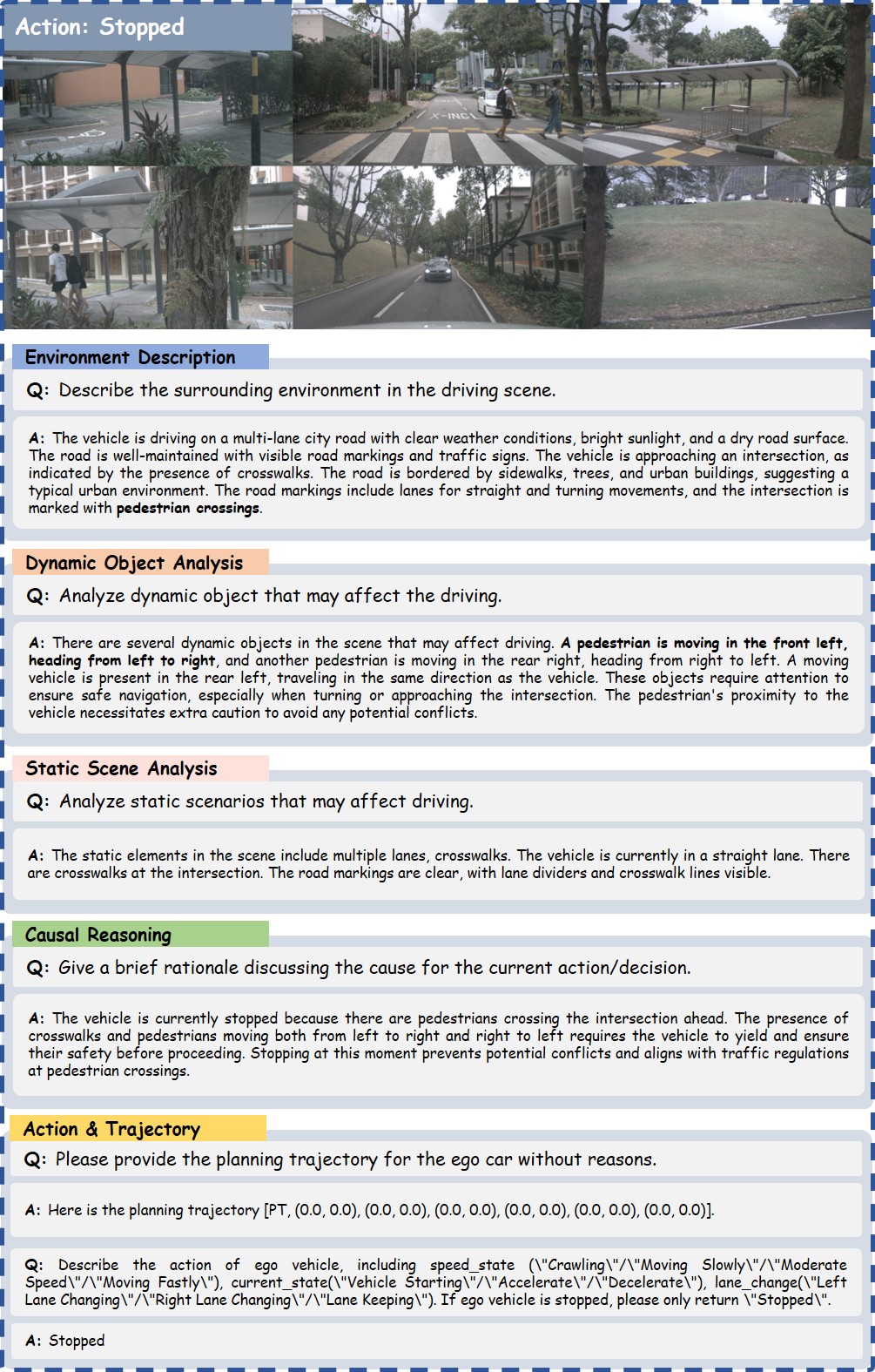}
    \caption{A stop VQA results of {\model}-Agent on the nuScenes dataset.}
    \label{fig:pred-stop}
\end{figure*}

\begin{figure*}    
\centering
    \includegraphics[width=0.85\linewidth]{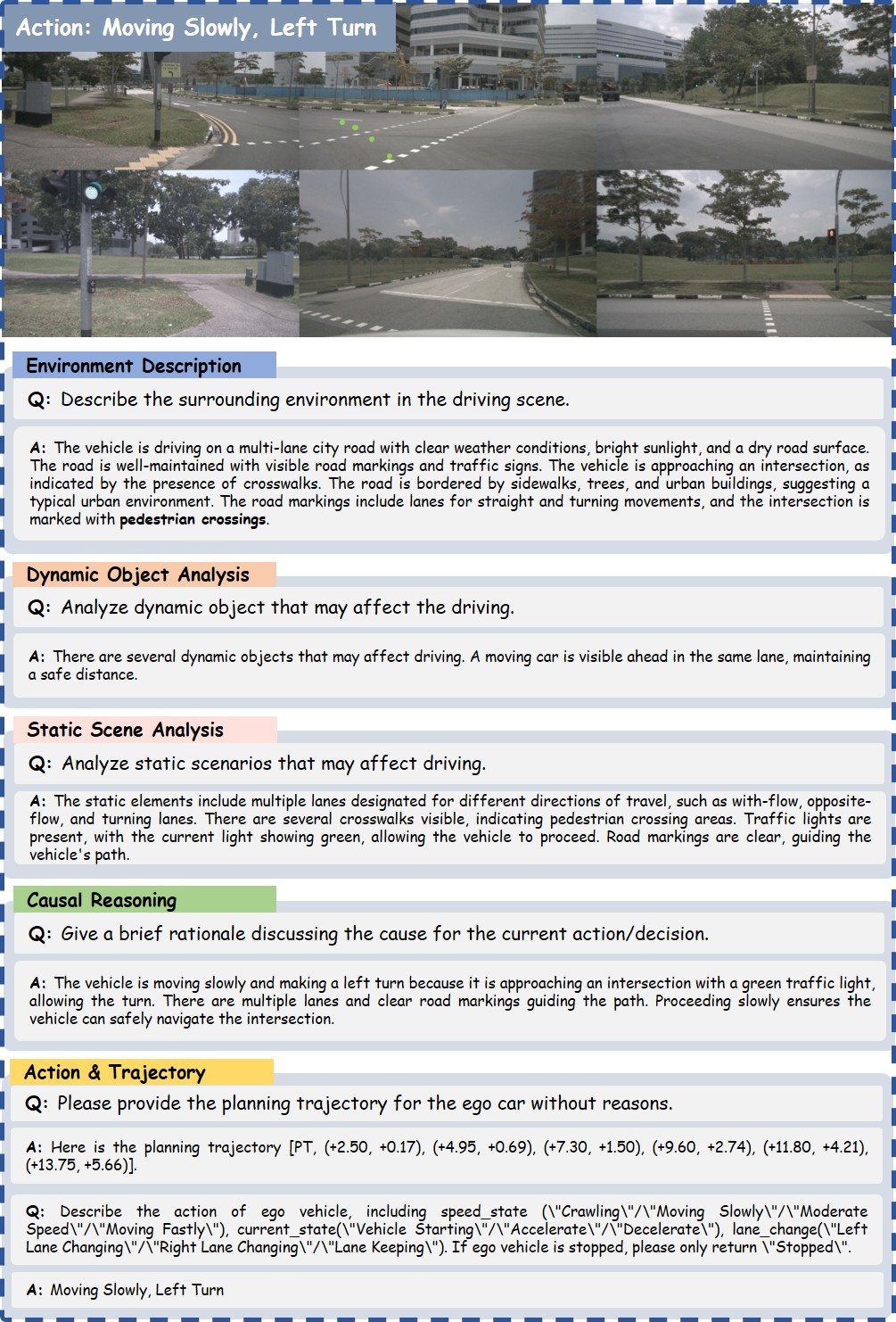}
    \caption{A left turn VQA results of {\model}-Agent on the nuScenes dataset. The green refers to the predicted trajectory.}
    \label{fig:pred-left-turn}
\end{figure*}

\clearpage
\section{{\model}-Agent}
The training of {\model}-Agent follows a two-stage curriculum designed to first align modalities and then fine-tune for the end-to-end driving task. This strategy effectively leverages the frozen LLM core while ensuring stable convergence. 

\begin{figure}[h]
    \centering
    \includegraphics[width=1.0\linewidth,  keepaspectratio]{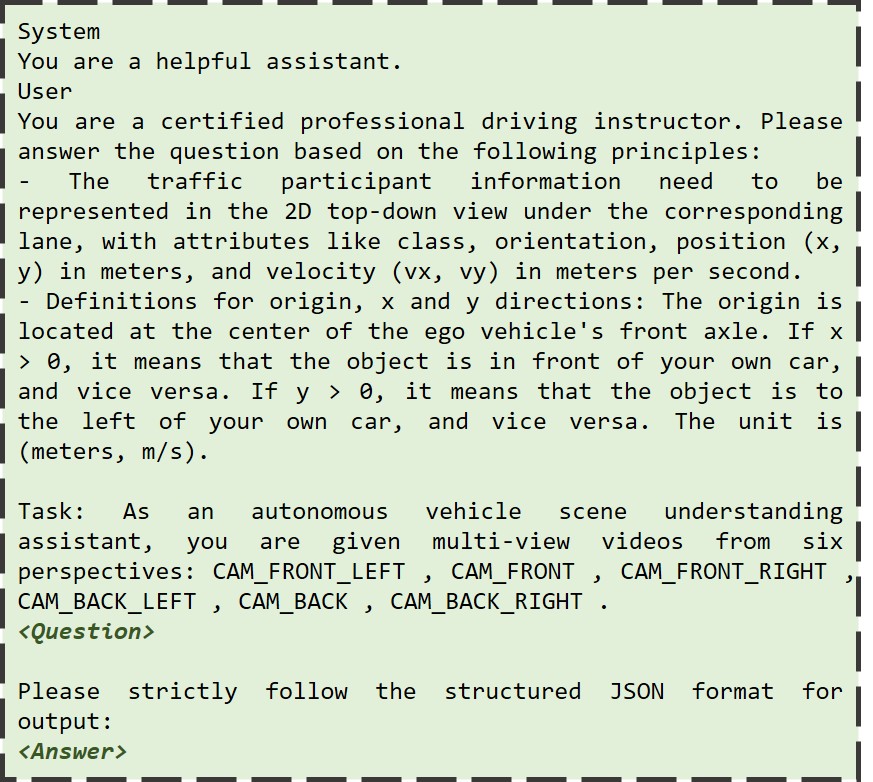}
    \caption{Prompt template for {\model}-Agent.}
    \label{fig:template2}
\end{figure}

\textbf{Vision-Language Alignment.} The initial stage trains the vision encoder and a set of lightweight adapters to produce meaningful features for the frozen LLM. Using our {\model}-Data, the model learns to generate the ground-truth narrative explanation from multi-view video clips and a structured prompt, the template for which is detailed in Fig. \ref{fig:template2}. 

\textbf{End-to-End VLA Fine-tuning.} In the second stage, we fine-tune the agent for the full vision-language-action task. The vision encoder and adapters remain trainable, while the LLM core stays frozen. The model is optimized to jointly predict both the narrative and the future trajectory waypoints. 

\section{Experiments}

\subsection{Datasets and Evaluation Metrics}
The proposed {\model}-Data encompasses open-loop planning, closed-loop planning, and VQA tasks. We evaluate {\model}-Agent on the open-loop planning task using the nuScenes benchmark, measuring planning performance via L2 displacement errors at 1, 2, and 3 seconds, along with the average collision rate (CR) and intersection rate (IR). To assess performance in more realistic, interactive scenarios, we evaluate closed-loop planning on the Bench2Drive dataset using Driving Score, Success Rate, Efficiency, and Comfortness. To further probe the model’s scene understanding and reasoning capabilities, we assess its performance on our driving VQA dataset after instruction tuning. For VQA evaluation, we adopt standard metrics: CIDEr (CI-r), BLEU-1 (BL-1), BLEU-4 (BL-4), METEOR (ME-R), ROUGE-L (RO-L), Precision, and Recall, ensuring a comprehensive analysis of language understanding and multimodal alignment.

\subsection{Ablation Study}
\subsubsection{Generalizability on OmniDrive Dataset.}

Table \ref{tab:ab} presents a comprehensive evaluation of {\model}-Agent against state-of-the-art multimodal models on the OmniDrive dataset. Our model establishes new benchmarks across all evaluation metrics, achieving a CIDEr score of 86.67, representing a 3.0\% improvement over the previous best model LLaVa-next 72B. The BLEU-1 score of 36.40 demonstrates an even more substantial 7.1\% gain, while the BLEU-4 score shows a remarkable 42.4\% improvement.
The performance advantage is particularly notable when comparing models of similar scale. Our model outperforms Qwen2.5VL 3B by 18.1\% in CIDEr and 23.0\% in BLEU-1, despite identical model sizes. More significantly, our solution surpasses much larger 72B parameter models, exceeding Qwen2.5VL 72B by 7.7\% in CIDEr and LLaVa-next 72B by 3.0\% in the same metric.
This comprehensive performance advantage demonstrates that {\model}-Agent's architectural innovations deliver superior results regardless of model scale, establishing new state-of-the-art performance for autonomous driving applications.

\begin{table}
    \centering
    \renewcommand{\arraystretch}{1.2}
    \scalebox{0.89}{
    \begin{tabular}{l|ccccc}
    \hline
        \textbf{Model} &  \textbf{CI-r} & \textbf{BL-1} & \textbf{BL-4} & \textbf{ME-R} & \textbf{RO-L}\\
        \hline
        Qwen2.5VL 72B & 80.45 & 32.96 & 4.46 & 40.41 & 23.56 \\
        Qwen2.5VL 32B & 76.13 &25.86 &2.82 & 29.36& 20.77  \\
       Qwen2.5VL 7B & 76.93& 32.46&3.95 &44.46 & 23.11 \\
        Qwen2.5VL 3B & 73.41  &29.59& 3.50 &46.70 & 23.48 \\
        LLava-next 72B & 84.14 & 33.98 & 5.28 &42.44 & 23.87 \\
        
        \hline
        {\model}-Agent & \textbf{86.67}  & \textbf{36.40} & \textbf{7.52} & \textbf{47.58}  & \textbf{27.87} \\
        \hline
    \end{tabular}
    }
    \caption{Performance comparison on the OmniDrive dataset \cite{wang2025omnidrive}. {\model}-Agent outperforms prior works in all metrics.}
    \label{tab:ab}
\end{table}

\begin{figure*}[t!]
\centering

\includegraphics[width=1.0\linewidth]{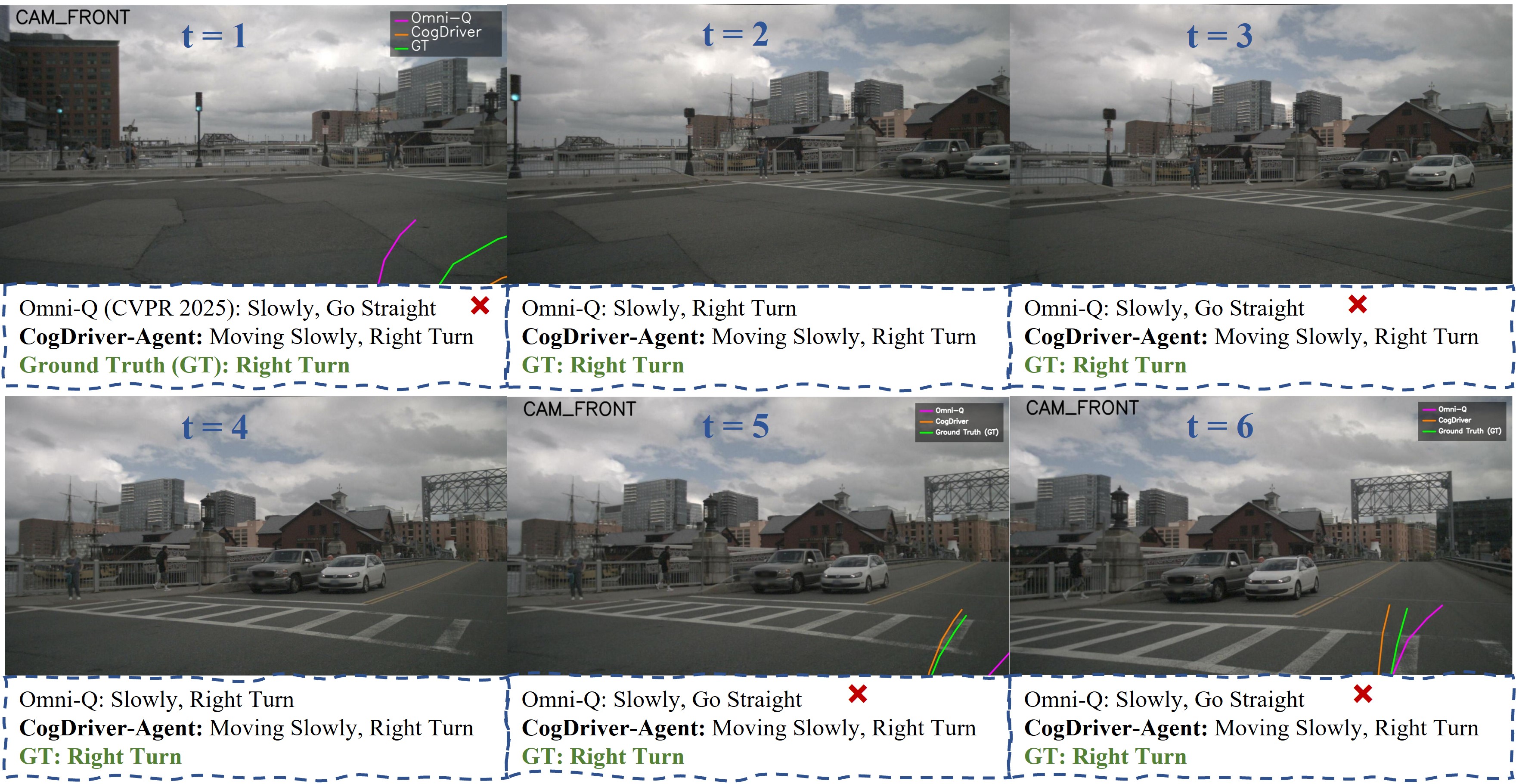}

\caption{Qualitative comparison of decision coherence.}
\label{fig:quli_comp}

\end{figure*}

\subsection{Qualitative Results}

Fig.~\ref{fig:pred-stop} and Fig.~\ref{fig:pred-left-turn} showcase the VQA results of {\model}-Agent on the nuScenes dataset for two distinct urban intersection scenarios. In Fig.~\ref{fig:pred-stop}, the ego vehicle identifies the presence of pedestrians crossing from both directions and makes an interpretable, safety-oriented decision to stop, demonstrating strong scene understanding and adherence to traffic rules. Fig.~\ref{fig:pred-left-turn} presents a left turn scenario where the agent recognizes a green traffic light and clear road markings, and cautiously executes a slow left turn while monitoring a vehicle ahead. Together, these examples highlight the model’s ability to integrate static and dynamic scene information, perform causal reasoning, and generate reliable, context-aware actions in complex real-world traffic situations.

Fig.~\ref{fig:quli_comp} visualizes the causal alignment between decisions and trajectories. While the baseline (Omni-Q) suffers from severe instability, erratically flipping between ``Go Straight" and ``Right Turn", CogDriver-Agent demonstrates robust Cognitive Inertia. Our {\model}-Agent maintains a consistent, long-term driving intent that aligns perfectly with the Ground Truth, proving its ability to generate stable, causally grounded plans despite visual fluctuations.

\end{document}